\pdfoutput=1

\documentclass[11pt]{article}

\usepackage[final]{acl}

\usepackage{times}
\usepackage{latexsym}

\usepackage[T1]{fontenc}

\usepackage[utf8]{inputenc}

\usepackage{microtype}

\usepackage{inconsolata}

\usepackage{graphicx}

\usepackage{colortbl}
\usepackage{xcolor}
\usepackage{booktabs}
\usepackage{multirow}
\usepackage{amsmath}
\usepackage{cleveref} 
\usepackage{subcaption}
\usepackage{graphics}
\usepackage{xspace}
\usepackage{float}
\usepackage{CJK}

\usepackage{tipa}
\usepackage{longtable}

\usepackage{algorithm}
\usepackage{algpseudocode}
\usepackage{amsmath}

\usepackage{dingbat}

\usepackage[table]{xcolor}   

\definecolor{green}{HTML}{e9f9c0}
\definecolor{red}{HTML}{f9c0c0}
\definecolor{yellow}{HTML}{f6ea94}

\title{GKnow: Measuring the Entanglement of Gender Bias and Factual Gender}

\author{Leonor Veloso \and Hinrich Schütze \\
         Center for Information and Language Processing, LMU Munich \\ Munich Center for Machine Learning (MCML) \\ \texttt{lveloso@cis.lmu.de}}

\newcounter{notecounter}

\newcommand{\enoteson}{\long\gdef\enote##1##2{{
\stepcounter{notecounter}
{\large\bf
\hspace{0cm}\arabic{notecounter} $<<<$ ##1: ##2
$>>>$\hspace{1cm}}}}}

\enoteson

\def\gknow{GKnow\xspace}

\begin{document}
\maketitle
\begin{abstract}

Recent works have analyzed the impact of individual components of neural networks on gendered predictions, often with a focus on mitigating gender bias. However, mechanistic interpretations of gender tend to (i) focus on a very specific gender-related task, such as gendered pronoun prediction, or (ii) fail to distinguish between the production of \textit{factually gendered outputs} (the correct assumption of gender given a word that carries gender as a semantic property) and \textit{gender biased outputs} (based on a stereotype). To address these issues, we curate \gknow, a benchmark to assess gender knowledge and gender bias in language models across different types of gender-related predictions. \gknow allows us to identify and analyze circuits and individual neurons responsible for gendered predictions. We test the impact of neuron ablation on benchmarks for disentangling stereotypical and factual gender (DiFair and the test set of \gknow), as well as StereoSet. Results show that gender bias and factual gender are
severely entangled on the level of both circuits and neurons, entailing that ablation is an unreliable debiasing method. Furthermore, we show that
benchmarks for evaluating gender bias can hide the decrease
in factual gender knowledge that accompanies neuron
ablation. We curate \gknow as a contribution to the
continuous development of robust gender bias benchmarks.


\end{abstract}

\section{Introduction}

Thanks to recent developments in the field of \textit{mechanistic
interpretability}, we have  a growing understanding
of why and how a large language model
(LLM) produces a certain
output \citep{conmy2023towards}. Mechanistic analysis
methods have been applied to the production of socially
biased outputs. For the specific case of gender bias,
techniques such as causal mediation
analysis \citep{vig2020causal,cai2024locating,
chintam2023identifying} have been employed to locate
relevant components for the production of gender biased
outputs. More interpretable debiasing techniques, such as
probing \citep{orgad2022gender}, concept
erasure \citep{belrose2024leace}, and neuron ablation \citep{liu2024devil,kavuri2025neat,chandna2025dissecting} are gaining traction due to
their lack of reliance on costly curation of datasets and
expensive fine-tuning.

An often overlooked side effect of gender debiasing is
damage to what can be  referred to as the
model's \textit{factual gender
knowledge} \citep{zakizadeh2023difair}, i.e., the lexical
correspondence between nouns that possess the semantic
property of \textit{female} or \textit{male} and their
respective pronouns or determiners (e.g., \textit{woman/she} and \textit{man/he}).
This phenomenon is a symptom of the entanglement
of gender bias and factual gender signals in the inner model
representations. Previous work has focused on the
development of debiasing methods that keep the factual
gender signal \citep{limisiewicz2022don,limisiewiczdebiasing}, but the
mechanistic story behind this entanglement on a circuit level remains mostly unexplained.

Our contributions\footnote{\url{https://github.com/leonorv/gknow}} are as follows:

(i) We curate \gknow, closing the gap in available benchmarks for evaluating the entanglement of gender bias and gender knowledge in English, for different types of gender-related tasks;

(ii) Employing the state-of-the-art circuit analysis technique EAP-IG, we find evidence of entanglement between gender bias and factual gender on a circuit-level. This circuit-level analysis is described in \Cref{sec:Circuit Analysis}.

(iii) We employ the interpretability/debiasing technique of
neuron ablation for Llama-3.1-8b and Olmo-7b on the test set of \gknow, StereoSet \citep{nadeem2020StereoSet}, and DiFair \citep{zakizadeh2023difair},
a benchmark for assessing the entanglement of gender bias and factual gender. We detail our neuron-level analysis and experiments in \Cref{sec:Neuron-Level Analysis}.

We observe that circuits responsible for gender bias and
factual gender knowledge have a high overlap, and a low
degree of separation when measuring cross-task faithfulness
(i.e., the ability of one circuit to solve another's
task \citep{hanna2025formal}). Since this entanglement is
also present at the level of
neurons, simple ablation of the most relevant gender bias neurons causes a sharp decrease in the model's linguistic ability to correctly predict factual gender. However, we also find that ablation of stereotypical gender neurons has results that can be interpreted as positive in biased settings, masking the decrease in factual gender knowledge. This highlights the need to develop resilient gender bias evaluation datasets that take factual gender into account. Based on these findings, we caution against unfiltered neuron ablation as the sole method of gender debiasing.

\section{Related Work}

\subsection{Linguistic Gender in English}

Although English lacks grammatical gender, it possesses \textit{lexically gendered nouns} and \textit{socially gendered nouns} \citep{motschenbacher2016discursive}. Lexical and social gender are linguistic gender categories: lexical gender refers to the semantic
association between a noun and its gender (\textit{woman/she} and \textit{man/he}), while social gender
conveys stereotypical femaleness or maleness and can differ across time and cultures (\textit{nurse/she} and \textit{pilot/he}). Following other NLP works \citep{zakizadeh2023difair,limisiewicz2022don}, we will refer to \textit{social gender} as \textit{stereotypical gender}, and to \textit{lexical gender} as \textit{factual gender}.


\subsection{Circuit Discovery and Neuron Attribution}

A \textit{circuit} is defined in the field of mechanistic
interpretability as a computational subgraph with distinct
functionality \citep{wang2022interpretability}. Circuits can
be interpreted at different levels of granularity and across
different components. The nodes of the subgraph represent
model components, including neurons, attention heads, and
embeddings, while the edges symbolize interactions between
these components, such as residual connections, projections,
or attention
mechanisms \citep{yao2024knowledge,conmy2023towards}. Previous
works have identified the circuits relevant for the
production of gendered outputs, using methods such as causal
mediation analysis and activation/attribution
patching \citep{chintam2023identifying,
mathwin2023identifying,chandna2025dissecting}. Recent
efforts at benchmarking and standardizing mechanistic
interpretability methods have led to a predominant interest
in edge-based
circuits \citep{mueller2025mib,ferrando2024information,hanna2024have}.
Building on this work, we apply circuit analysis to interpret gender.

An adjacent form of interpretability work comes in the form of neuron attribution -- i.e., identifying task-relevant hidden FFN neurons. Several works have pinpointed the importance of neurons in storing knowledge \citep{dai2021knowledge,geva2021transformer,yu2025understanding}. Different methods have been proposed to calculate the importance of a neuron to a prediction, from gradient-based \citep{dai2021knowledge} to attribution-based (or static) methods \citep{yu2024neuron}. A small subset of neurons can play a critical role in several model capabilities, such as language competence \citep{duan2024unveiling}, factual knowledge \citep{dai2021knowledge,chen2024journey}, and linguistic phenomena \citep{niu2024does}. These works also show that simple ablation of subsets of relevant neurons impacts model behavior. Of note to the present  work, \citet{liu2024devil,kavuri2025neat,yu2025understanding} identify neurons relevant for the production of socially and gender biased outputs (respectively), and evaluate neuron ablation as a debiasing technique. In our work, we use neuron ablation both to test  for gender debiasing and to further our understanding of gender circuits.

\subsection{Impact of Gender Debiasing on Factual Gender}

The entanglement of gender bias and factual gender can be
observed in model representations \citep{limisiewicz2022don}
and feature vectors \citep{dunefsky2024observable}. This
entanglement raises the concern that gender debiasing
methods might negatively affect a model's knowledge of
factual gender. \citet{zakizadeh2023difair} introduce a
language modeling dataset that aims to measure performance
on gendered instances, and find that bias mitigation methods
can impair factual gender information. Other studies have
introduced embedding debiasing techniques that focus on
preservation of factual gender
information \citep{bolukbasi2016man}. \citet{zhao2018learning} exempt lexically
gendered words from debiasing. Other work  removes
stereotypically gendered signals while keeping factually
gendered signals via
probing \citep{limisiewicz2022don}
or
projecting the original embeddings to a
debiased space \citep{kaneko2019gender}.

\section{Experimental Setup}
\label{sec:Experimental Setup}

\subsection{Datasets}
\label{subsec:Datasets}

\subsubsection{GKnow}
\label{subsubsec:Gknow}

We curate \textit{GKnow}, a benchmark for evaluation of gender-related tasks in autoregressive models. Gknow entries are categorized by a type of gender \textit{assumption} (in the form of the subject) and a gendered \textit{prediction} (the expected output of the prompt). For example, in the sentence \texttt{The woman is nice, isn't [MASK]}, a model predicts \texttt{she}. In this case, \texttt{she} is a form of gendered pronoun prediction that is based on the lexical gender (the assumption) of the subject \texttt{woman}. \Cref{tab:gknow_example} depicts two entries of the test set of \gknow.

\Cref{tab:all_prompt_categories} provides examples for each gender-related category. We selected these categories to account for different ways linguistic gender can manifest itself in the English language. Therefore, our categories of possible gendered subjects/expected outputs include pronouns \citep{mathwin2023identifying,dunefsky2024observable}, indicators of gender \citep{liu2024devil,hernandez2023linearity}, names \citep{mathwin2023identifying}, lexically gendered nouns, and stereotypically gendered nouns \citep{yu2025understanding,bartl2022inferring}. The gender-related terms and prompt templates we use for the construction of \gknow are partially derived from datasets/prompts used in other works, concerned either with mechanistic interpretations of gender or the entanglement of bias and factual gender. Our stereotypical terms can be occupations or traits/adjectives, and our lexically gendered terms can be semantically related to \textit{family}, \textit{occupations}, \textit{religion}, and more.

\gknow prompts that include a stereotypical assumption or entail a stereotypical
prediction are considered \textit{stereotypical prompts}, as
opposed to \textit{factual} prompts. This allows for the usage of \gknow as a benchmark to assess the entanglement of gender bias and factual gender, akin to DiFair \citep{zakizadeh2023difair}.

In summary, \gknow can be
used as (i) a resource to perform mechanistic analyses of
gender across different gender-related tasks; and (ii) a benchmark for testing autoregressive models'
entanglement of stereotypical and factual gender
knowledge. The full version of \gknow has 91,490 examples (in the present work, we use a smaller version of 6992 examples),
serving as an extensive resource for mechanistic analyses across gender-related tasks and subcategories of gender-related terms. See \Cref{app:gknow_details}
for details regarding dataset curation.

\begin{table}[h]
  \centering
  \small
  \begin{tabular}{lp{4.5cm}}
  \toprule
  \textbf{Key} & \textbf{Value} \\
  \midrule
  prompt & The female person wished that \\
  subject & female person \\
  expected\_output & she \\
  gender & feminine \\
  id & 18 \\
  \midrule
  \midrule
  prompt & The gender of the nurturing person is? Answer: \\
  subject & nurturing person \\
  expected\_output & Female \\
  stereo\_category & adjective \\
  gender & feminine \\
  id & 9985 \\

  \bottomrule
  \end{tabular}
  \caption{Example entries from the \textit{pronoun prediction based on gender} (top) and \textit{gender prediction based on stereotypical} subsets of \gknow (bottom). \textit{Stereotypical} subsets are enriched with metadata regarding the stereotype category (occupational or adjective/trait based).}
  \label{tab:gknow_example}
  \end{table}

{

\renewcommand{\arraystretch}{1.5}
\setlength{\tabcolsep}{4pt}

\newcolumntype{C}[1]{>{\centering\arraybackslash}m{#1}}
\newcommand{\cuteCell}[2]{\cellcolor{#1}\sffamily #2}
\definecolor{babyblue}{HTML}{DCEEFB}
\definecolor{pastelpink}{HTML}{FDE2E2}

\begin{table*}[ht]
\centering
\resizebox{\textwidth}{!}{%
\begin{tabular}{ll C{3.5cm} C{3.5cm} C{3.5cm} C{3.5cm} C{3.5cm}}
\toprule
& & \multicolumn{5}{c}{\textbf{Assumption}} \\
\cmidrule(lr){3-7}
& & \textbf{Gendered Pronoun} & \textbf{Gender} & \textbf{Gendered Name} & \textbf{Factually Gendered Term} & \textbf{Stereotypically Gendered Term} \\
\midrule

& \textbf{Gendered Pronoun} &
& \cuteCell{babyblue}{The woman is nice, isn’t [she]} 
& \cuteCell{babyblue}{Mary is nice, isn’t [she]} 
& \cuteCell{babyblue}{The sister is nice, isn’t [she]} 
& \cuteCell{pastelpink}{The nurse is nice, isn’t [she]} \\

& \textbf{Gender} &
\cuteCell{babyblue}{She identifies as [female]} 
& 
& \cuteCell{babyblue}{Mary identifies as [female]} 
& \cuteCell{babyblue}{The sister identifies as [female]} 
& \cuteCell{pastelpink}{The nurse identifies as [female]} \\

& \textbf{Gendered Name} &
\cuteCell{babyblue}{Her name is [Mary]} 
& \cuteCell{babyblue}{The woman is called [Mary]} 
& 
& \cuteCell{babyblue}{The sister’s name is [Mary]} 
& \cuteCell{pastelpink}{The nurse’s name is [Mary]} \\

\multirow{-5}{*}{\textbf{Prediction}} & \textbf{Factually Gendered Term} &
\cuteCell{babyblue}{She is a [mother]} 
& \cuteCell{babyblue}{The woman is the [sister]} 
& \cuteCell{babyblue}{Mary is a [mother]} 
& 
& \cuteCell{pastelpink}{The nurse is a [mother]} \\

& \textbf{Stereotypically Gendered Term} &
\cuteCell{pastelpink}{She is a [nurse]} 
& \cuteCell{pastelpink}{The woman is a [nurse]} 
& \cuteCell{pastelpink}{Mary is a [nurse]} 
& \cuteCell{pastelpink}{The sister is a [nurse]} 
& \\
\bottomrule
\end{tabular}
}
\vspace{.75em}
\caption{Categories of gender-related prompts used in
    the \gknow dataset.
The prediction to be tested appears in square brackets. The
    prompt before the square brackets corresponds to the assumption the prediction is based on.
Blue cells represent prompt types that focus on factual
    gender, while red cells represent prompts that focus on
    stereotypical gender/bias analysis. Over the course of this work, we refer to \gknow subsets according to their prediction and assumption -- for instance, \texttt{pronoun\_prediction\_based\_on\_stereo} would refer to the top right prompts; \texttt{gender\_prediction\_based\_on\_name} would refer to prompts such as ``Mary identifies as [female]'', etc. We are aware that
    the assumption of gender when given a name can also be problematic or inaccurate, but we follow current NLP literature  and treat names as manifestations of factual gender \citep{mathwin2023identifying}. ``Gender'' refers to explicit gender indicators (``male'', ``female'', ``man'', ``woman'').}
  \label{tab:all_prompt_categories}
\end{table*}

}

\subsubsection{StereoSet \& DiFair}



To assess generalization of the effects of neuron ablation identified with \gknow across other benchmarks, we evaluate it as a debiasing method on StereoSet \citep{nadeem2020StereoSet} and DiFair \citep{zakizadeh2023difair}. Although StereoSet has seen extensive criticism within the
bias and fairness
literature \citep{blodgett2021stereotyping,orgad2022choose},
it has been used for evaluating
neuron ablation \citep{liu2024devil,yu2025understanding}. Motivated by our usage of relatively small models (and similarly to \citeauthor{liu2024devil}), we use the \textit{intrasentence} subset of StereoSet, reducing the scope of the evaluation to sentence-level reasoning. Each intrasentence entry of StereoSet contains a sentence and three potential completions -- stereotypical, anti-stereotypical, and a contextually unrelated term.

DiFair \citep{zakizadeh2023difair} is a benchmark for the assessment of gender bias and gender knowledge. DiFair is a manually annotated dataset, constructed from text from the English Wikipedia and Reddit. It is split into \textit{gender-neutral} sentences, which have no gender cues, and \textit{gender-specific} sentences, which contain gender cues in the form of pronouns, names, historical and biological references. An entry of DiFair does not specify an expected completion token/term, but the dataset suite contains lists of feminine and masculine terms as possible gendered completions. 


In the interest of decoupling metrics (further details
in \ref{subsec:Metrics}), we use only entries of these
evaluation datasets where the masked token is the final
token. This yields 104 examples from the ``gender'' category of StereoSet, as well as 63 ``gender-neutral'' and 49 ``gender-specific'' examples from DiFair.

\subsection{Metrics}
\label{subsec:Metrics}

The extensive literature on gender bias has led to the introduction of many benchmarks. In order to decouple metrics from their respective datasets/tasks and focus on extrinsic metrics \citep{orgad2022choose}, we report the following metrics:

\begin{itemize}
  \item $P_{exp}$, the probability of the expected token: \texttt{[she]} (resp.\ \texttt{[he]}) for a factually or stereotypically feminine (resp.\ masculine) sentence. Requires the benchmark to specify an expected/sterotypical output (note this is not the case for DiFair). 
  \item $P_{opp}$, probability of the opposite binary gendered token (i.e., the one that is not expected). For \gknow, we augment the \texttt{gender\_prediction} and \texttt{pronoun\_prediction} subsets with the binary-gendered token opposite to the expected token;
  \item $P_{other}$, probability of outputting a third
  token. For \gknow, we augment
  the \texttt{gender\_prediction}
  and \texttt{pronoun\_prediction} subsets with a neutral
  token.\footnote{\texttt{they} for pronoun
  prediction; \texttt{person} for gender prediction.} In
  StereoSet, this corresponds to the contexually unrelated
  completion.
In \citet{orgad2022choose}'s categorization of gender bias metrics,
$P_{exp}$, $P_{opp}$, $P_{other}$  relate to the models's prediction on target words;
  \item $\%exp$, percentage of examples where the model prefers the expected binary-gendered token in lieu of others,
  \item $\%opp$, percentage of examples where the model prefers the opposite binary-gendered token,
  \item $\%other$, percentage of examples where the model
  prefers a third token.
In \citet{orgad2022choose}'s categorization of gender bias metrics,
$\%exp$, $\%opp$, $\%other$, relate to the model's preference.
  \item $\Delta$, probability gap between the masculine and feminine outputs (for DiFair and GKnow). Note that, since the anti-stereotypical completions of StereoSet are not necessarily associated with the opposite gender of the prompt's subject, we do not calculate this metric for it. Within the categorization of \citep{orgad2022choose}, this falls under the prediction gap category. Can be interpreted as a metric of neutrality or confidence in the model's binary-gendered decision.
\end{itemize}

While we report these metrics in the interest of decoupling benchmarks and metrics and analyzing the probability distribution after neuron
ablation, these can be used for calculating more fine-grained
metrics if so desired, such as StereoSet's ICAT
score \citep{nadeem2020StereoSet} or DiFair's
GIS \citep{zakizadeh2023difair}. Note that the content and
structure of the datasets impacts the applicable metrics --
DiFair does not specify an expected or stereotypical output
for each example (rendering all metrics unapplicable except
$\Delta$, where we use the original authors' formulation for
probability gap -- between the maximum probability gendered
masculine and feminine tokens, which are retrieved from
lists of binary-gendered words). Due to these differences in
content, structure, and conceptualization of bias, direct
comparison of results across datasets can be
misleading \citep{blodgett2021stereotyping}. This is
especially the case for $P_{other}$ and $\%other$:
StereoSet's ``other'' completion is an invalid completion, while GKnow's ``other'' completion can create a grammatically correct gender-neutral sentence.




\subsection{Models}
\label{subsec:Models}

We conduct experiments with
Olmo-7b \citep{groeneveld2024olmo} and
Llama-3.1-8b \citep{touvron2023llama}, two decoder-style
transformer \citep{vaswani2017attention}
models. See \Cref{sec:appendixA} for architectural details.

\section{Circuit Analysis}
\label{sec:Circuit Analysis}

We leverage EAP-IG \citep{hanna2024have}, as the highest performing method on the circuit analysis track of MIB \citep{mueller2025mib}. EAP-IG is designed as a combination of edge attribution patching and integrated gradients \citep{sundararajan2017axiomatic}. Formally, if $u$ and $v$ are nodes in a model's computational graph, $m$ is the number of steps used to approximate the integral, $z$ is a sequence of token embeddings for one input, and $z'$ is the token embeddings of the distinct, baseline input, the EAP-IG score of the edge $(u, v)$ is:

\[
(z'_u - z_u) \frac{1}{m} \sum_{k=1}^{m} 
\frac{\partial L \big( z' + \frac{k}{m}(z - z') \big)}{\partial z_v},
\]

where $z'_u$ and $z_u$ are the corrupted and clean activations. We set $m=5$, as suggested by \citep{hanna2024have}. Since EAP-IG relies on counterfactual prompts, we augmented the train split of \gknow to include corrupted prompts. In the main document we focus our analysis on the \texttt{pronoun\_prediction} and \texttt{gender\_prediction} sets, since these have a higher expected output probability. More details regarding EAP-IG, the data augmentation process for \gknow, and extra results can be found in \Cref{app:Circuit Analysis: Details}. Operating under the circuit analysis framework of \citep{hanna2024have}, we aim to find the smallest faithful circuits (recovering >=80\% of the model's behavior). With \gknow entries as input, we identify minimal faithful circuits for each data subset. The structure of GKnow allows us to observe the circuit-level differences between different types of gendered predictions, where we are especially interested in the entanglement of stereotypical and factual circuits (\texttt{based\_on\_stereo} subsets vs. others).

\begin{figure*}[]
  \centering
      \centering
    \includegraphics[width=\textwidth]{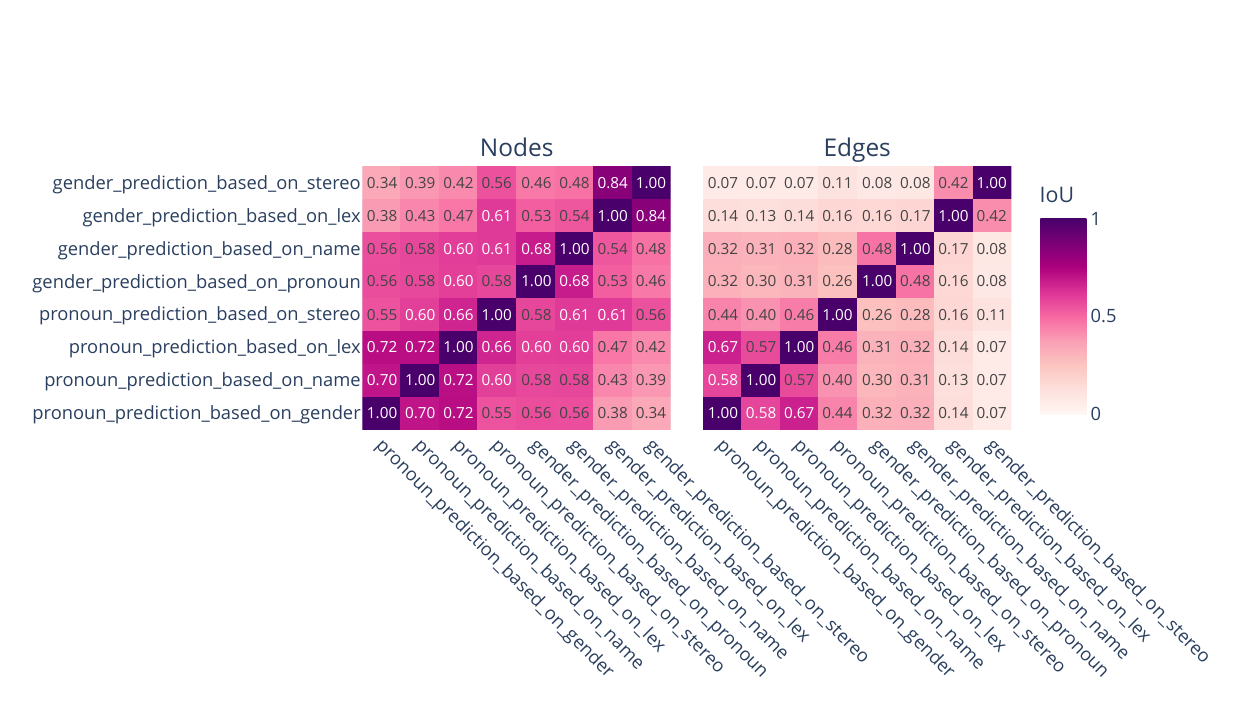}
  \caption{Edge and node intersection over union (Jaccard similarity) for minimal, faithful circuits in Llama.}
  \label{fig:llama_overlap}
\end{figure*}

\begin{figure}[]
  \centering
      \centering
    \includegraphics[width=\columnwidth]{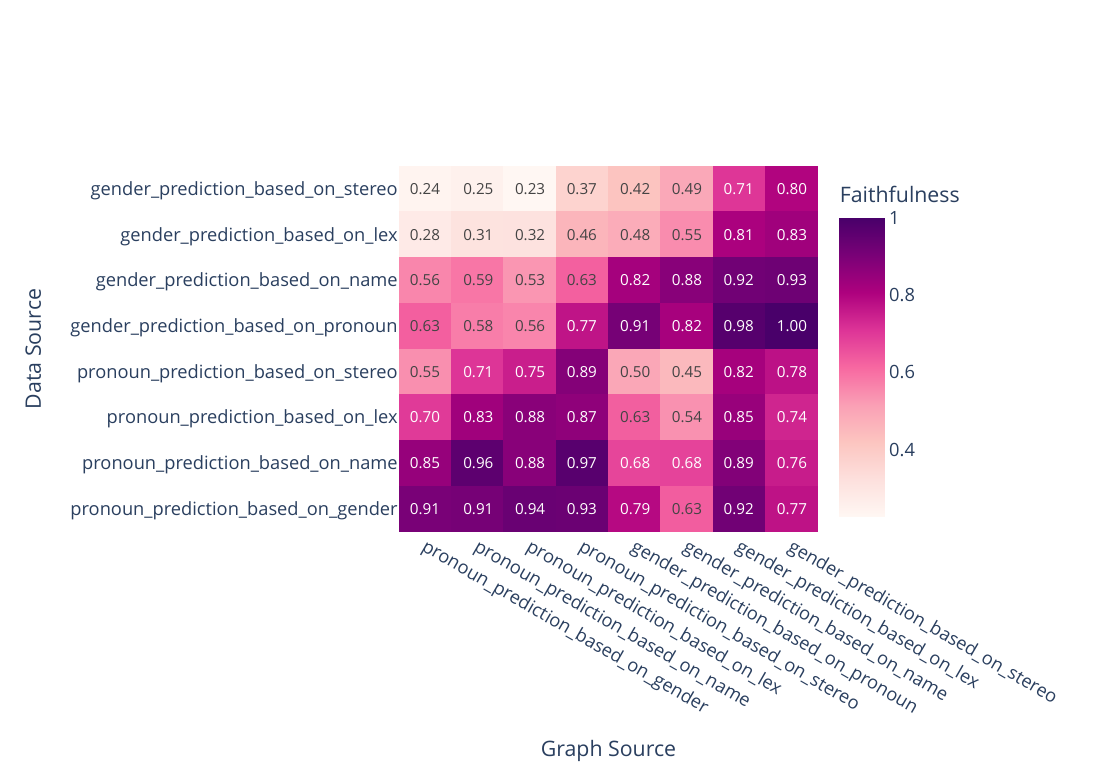}
  \caption{Cross-task faithfulness between the gendered tasks of GKnow, for Llama. Performance recovery is higher when a circuit (graph source) is applied to a task (data source) within the same type of prediction. Factual circuits are able to recover the majority of the performance of stereotypical tasks, and vice-versa: for example, the \texttt{gender\_prediction\_based\_on\_stereo} graph achieves a faithfulness of $1$ in the \texttt{gender\_prediction\_based\_on\_pronoun} subset.}
  \label{fig:llama_crosstask_faith}
\end{figure}



\begin{figure}[]
    \centering
    \begin{subfigure}{\columnwidth}
        \centering
        \includegraphics[width=\columnwidth]{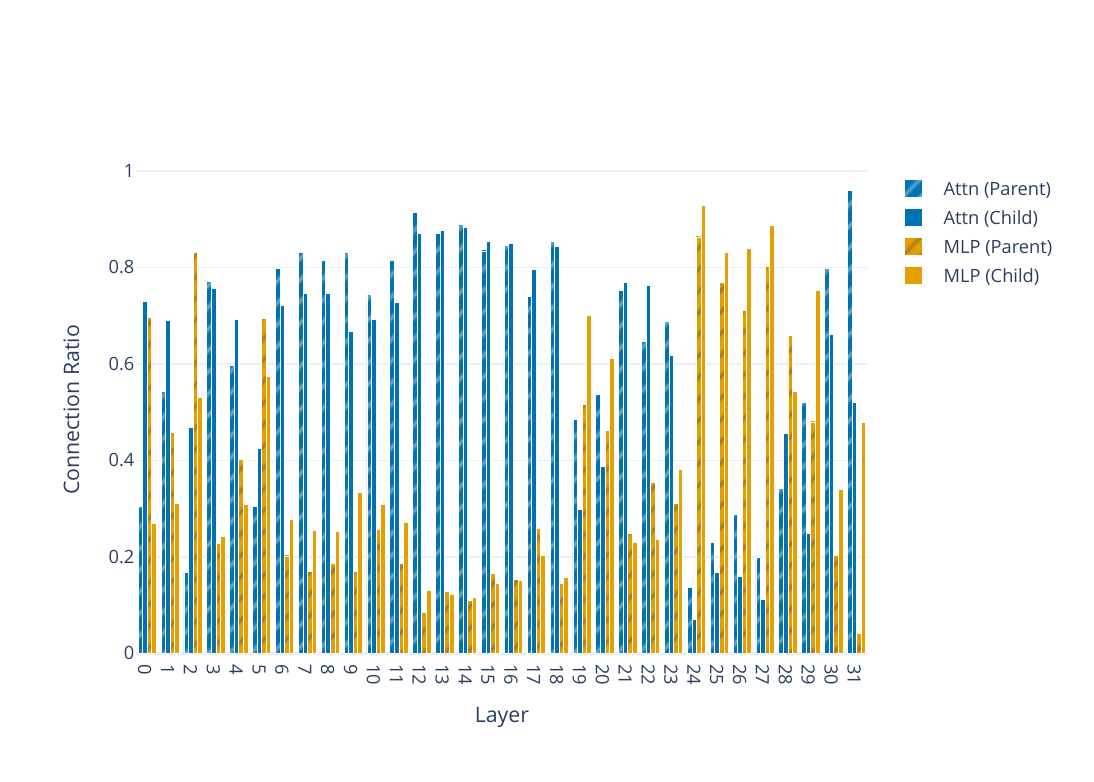}
        \label{fig:llama_localization}
    \end{subfigure}

    \vspace{-3em}

    \begin{subfigure}{\columnwidth}
        \centering
        \includegraphics[width=\columnwidth]{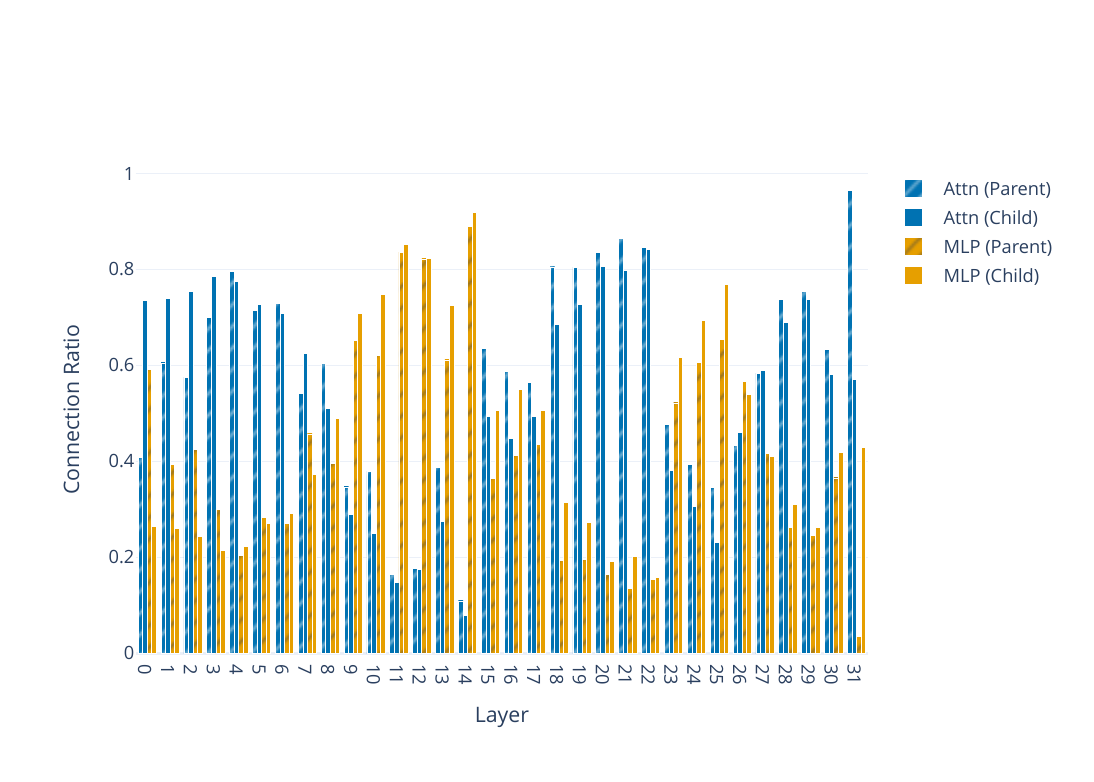}
        \label{fig:olmo_localization}
    \end{subfigure}

    \caption{Ratio of different types of connections within each layer, for Llama (top), and Olmo (bottom), for the intersection circuit of all gender-related tasks. Each edge has a parent and a child node, which can be MLPs or attention heads (the connection ratio of parents sum up to one, and similarly for the children). Circuits differ across models -- within the middle layers, Llama circuits have attention-centric dynamics, while Olmo's have predominantly connections between MLPs.}
    \label{fig:circuit_localization}
\end{figure}

To get an insight into the localization of important edges, we take the intersection circuit for all tasks, and compare it across models. As depicted in \Cref{fig:circuit_localization}, we calculate the connection ratio -- the ratio of different types of edge connections (to and from MLPs or attention heads) within each layer of the model -- finding that gender-related dynamics can differ. To compare circuits across gender-related tasks, we calculate their intersection over union (IoU) (\Cref{fig:llama_overlap}) and cross-task faithfulness -- i.e., the faithfulness achieved when using the circuit graph for one subset to solve the gender-related task of a different subset (\Cref{fig:llama_crosstask_faith}). Observing these results, we can have two main takeaways:




(i) Cross-task faithfulness (depicted in \Cref{fig:llama_crosstask_faith}) is higher within the
same type of prediction (e.g. average faithfulness of 77.2 when applying \texttt{gender\_prediction} circuits to \texttt{gender\_prediction} subsets, \textit{versus} 72.9 when applying \texttt{gender\_prediction} circuits to \texttt{pronoun\_prediction} subsets). However, some subtask circuits
perform better than others when applied to other subtasks
-- \texttt{based\_on\_name} tasks, whose prompts have a
gendered name as a subject, recover the least amount of
performance in other tasks. \texttt{based\_on\_lex} tasks,
which have a lexically gendered noun as a subject, recover
the highest amount of performance across tasks. We
hypothesize that, due to this subset being more diverse in
terms of subject (including family-related,
occupation-related, and miscellaneous gendered nouns), its
circuit has a higher chance to generalize.

(ii) Stereotypical (\texttt{based\_on\_stereo}) circuits are able to recover performance of factual subsets, and vice-versa. Notably, applying the \texttt{gender\_prediction\_based\_on\_stereo} circuit on the \texttt{gender\_prediction\_based\_on\_pronoun} subset achieves complete faithfulness (1.0). In a symmetric fashion, the \texttt{based\_on\_lex} subsets also achieve high faithfulness in their counterpart \texttt{based\_on\_stereo} subsets. This entails that factual and stereotypical gender are severely entangled on a circuit level.





\section{Neuron-Level Analysis}
\label{sec:Neuron-Level Analysis}

Motivated by the evidence for entanglement of stereotypical
and factual gender on a circuit-level, described
in \Cref{sec:Circuit Analysis}, we now focus on
neurons. Neurons are of special interest in the intersection of mechanistic interpretability and gender bias, since they can be human-interpretable and have been a target for ablation-based debiasing. However, we hypothesize that the circuit-level entanglement of stereotypical and factual gender extends to individual neurons, thus compromising the success of ablation-based debiasing methods. 

Recent works that report successful results with neuron-based ablation do not work under the framework of circuit analysis, where the focus lies on identifying important connections (edges) between model components, but rather work under the framework of neuron attribution \citep{kavuri2025neat,yu2025understanding,liu2024devil}. Therefore, for a fairer assessment of neuron ablation as a gender debiasing method, we leverage \textit{integrated gradients}, due to the overall popularity of gradient-based methods and the previous usage of IG-based methods in debiasing \citep{liu2024devil}.

Individual neurons were identified using the training prompt split of \gknow (see Section \ref{subsec:Datasets}) as input sentences. The \textit{integrated gradients} method \citep{dai2021knowledge} evaluates the contribution of each neuron to a prediction (formalization in \Cref{app:Integrated Gradients Formal Description}). 

\begin{table*}[ht]
\centering
\small
\resizebox{\linewidth}{!}{%
\begin{tabular}{p{0.2cm} l c c c c c c c}
\toprule
N & Dataset & $P_{exp}$ & $P_{opp}$ & $P_{other}$ & $\%exp$ & $\%opp$ & $\%other$ & $\Delta_{f,m}$ \\
\midrule
\multirow{5}{*}{0} & GKnow Stereo & 67.66 & 28.90 & 3.43 & 78.75 & 21.25 & 0.00 & 21.81  \\
 & GKnow Factual & 91.49 & 7.17 & 1.33 & 100.00 & 0.00 & 0.00 & 43.77 \\
 & StereoSet & 65.26 & 26.27 & 8.48 & 65.38 & 20.19 & 14.42 & - \\
 & DiFair Neutral & - & - & - & - & - & - & 6.48 \\
 & DiFair Specific & - & - & - & - & - & - & 45.57 \\

 \midrule

\multirow{5}{*}{10} & GKnow Stereo & 63.59 {\scriptsize\textcolor{red!70!black}{$\downarrow$4.07}} & 26.38 {\scriptsize\textcolor{red!70!black}{$\downarrow$2.52}} & 10.03 {\scriptsize\textcolor{green!60!black}{$\uparrow$6.59}} & 77.50 {\scriptsize\textcolor{red!70!black}{$\downarrow$1.25}} & 17.50 {\scriptsize\textcolor{red!70!black}{$\downarrow$3.75}} & 5.00 {\scriptsize\textcolor{green!60!black}{$\uparrow$5.00}} & 16.73 {\scriptsize\textcolor{red!70!black}{$\downarrow$5.08}} \\
 & GKnow Factual & 90.12 {\scriptsize\textcolor{red!70!black}{$\downarrow$1.37}} & 7.92 {\scriptsize\textcolor{green!60!black}{$\uparrow$0.75}} & 1.96 {\scriptsize\textcolor{green!60!black}{$\uparrow$0.63}} & 98.90 {\scriptsize\textcolor{red!70!black}{$\downarrow$1.10}} & 1.10 {\scriptsize\textcolor{green!60!black}{$\uparrow$1.10}} & 0.00 {\scriptsize\textcolor{gray!70!black}{$\rightarrow$ 0.00}} & 38.52 {\scriptsize\textcolor{red!70!black}{$\downarrow$5.27}} \\
 & StereoSet & 64.06{*} {\scriptsize\textcolor{red!70!black}{$\downarrow$1.20}} & 26.68{*} {\scriptsize\textcolor{green!60!black}{$\uparrow$0.41}} & 9.25 {\scriptsize\textcolor{green!60!black}{$\uparrow$0.77}} & 65.38 {\scriptsize\textcolor{gray!70!black}{$\rightarrow$ 0.00}} & 20.19 {\scriptsize\textcolor{gray!70!black}{$\rightarrow$ 0.00}} & 14.42 {\scriptsize\textcolor{gray!70!black}{$\rightarrow$ 0.00}} & - \\
 & DiFair Neutral & - & - & - & - & - & - & 5.84{*} {\scriptsize\textcolor{red!70!black}{$\downarrow$0.64}} \\
 & DiFair Specific & - & - & - & - & - & - & 40.31 {\scriptsize\textcolor{red!70!black}{$\downarrow$5.25}} \\

 \midrule

\multirow{5}{*}{50} & GKnow Stereo & 47.03 {\scriptsize\textcolor{red!70!black}{$\downarrow$20.64}} & 23.81 {\scriptsize\textcolor{red!70!black}{$\downarrow$5.09}} & 29.16 {\scriptsize\textcolor{green!60!black}{$\uparrow$25.73}} & 52.50 {\scriptsize\textcolor{red!70!black}{$\downarrow$26.25}} & 17.50 {\scriptsize\textcolor{red!70!black}{$\downarrow$3.75}} & 30.00 {\scriptsize\textcolor{green!60!black}{$\uparrow$30.00}} & 7.45{\scriptsize\textcolor{red!70!black}{$\downarrow$14.36}} \\
 & GKnow Factual & 79.86 {\scriptsize\textcolor{red!70!black}{$\downarrow$11.63}} & 12.38 {\scriptsize\textcolor{green!60!black}{$\uparrow$5.21}} & 7.77 {\scriptsize\textcolor{green!60!black}{$\uparrow$6.43}} & 89.56 {\scriptsize\textcolor{red!70!black}{$\downarrow$10.44}} & 5.49 {\scriptsize\textcolor{green!60!black}{$\uparrow$5.49}} & 4.95 {\scriptsize\textcolor{green!60!black}{$\uparrow$4.95}} & 18.03 {\scriptsize\textcolor{red!70!black}{$\downarrow$25.76}} \\
 & StereoSet & 62.60{*} {\scriptsize\textcolor{red!70!black}{$\downarrow$2.66}} & 27.60{*} {\scriptsize\textcolor{green!60!black}{$\uparrow$1.33}} & 9.80{*} {\scriptsize\textcolor{green!60!black}{$\uparrow$1.32}} & 60.58 {\scriptsize\textcolor{red!70!black}{$\downarrow$4.81}} & 23.08 {\scriptsize\textcolor{green!60!black}{$\uparrow$2.88}} & 16.35 {\scriptsize\textcolor{green!60!black}{$\uparrow$1.92}} & - \\
 & DiFair Neutral & - & - & - & - & - & - & 2.23 {\scriptsize\textcolor{red!70!black}{$\downarrow$4.25}} \\
 & DiFair Specific & - & - & - & - & - & - & 15.30 {\scriptsize\textcolor{red!70!black}{$\downarrow$30.26}}\\
\bottomrule
\toprule
N & Dataset & $P_{exp}$ & $P_{opp}$ & $P_{other}$ & $\%exp$ & $\%opp$ & $\%other$ & $\Delta_{f,m}$\\
\midrule
\multirow{5}{*}{0} & GKnow stereo & 63.66 & 26.62 & 9.72 & 77.50 & 16.25 & 6.25 & 17.09 \\
 & GKnow factual & 90.07 & 7.95 & 1.98 & 98.90 & 1.10 & 0.00 & 40.01 \\
 & StereoSet & 65.55 & 26.63 & 7.82 & 68.27 & 19.23 & 12.50 & - \\
  & DiFair Neutral & - & - & - & - & - & - & 9.63 \\
 & DiFair Specific & - & - & - & - & - & - & 48.26 \\

 \midrule

\multirow{5}{*}{10} & GKnow Stereo & 56.29 {\scriptsize\textcolor{red!70!black}{$\downarrow$7.37}} & 29.15{*} {\scriptsize\textcolor{green!60!black}{$\uparrow$2.53}} & 14.55 {\scriptsize\textcolor{green!60!black}{$\uparrow$4.83}} & 65.00 {\scriptsize\textcolor{red!70!black}{$\downarrow$12.50}} & 27.50 {\scriptsize\textcolor{green!60!black}{$\uparrow$11.25}} & 7.50 {\scriptsize\textcolor{green!60!black}{$\uparrow$1.25}} & 11.18 {\scriptsize\textcolor{red!70!black}{$\downarrow$5.91}} \\
 & GKnow Factual & 85.52 {\scriptsize\textcolor{red!70!black}{$\downarrow$4.55}} & 11.37 {\scriptsize\textcolor{green!60!black}{$\uparrow$3.42}} & 3.12 {\scriptsize\textcolor{green!60!black}{$\uparrow$1.13}} & 97.25 {\scriptsize\textcolor{red!70!black}{$\downarrow$1.65}} & 2.75 {\scriptsize\textcolor{green!60!black}{$\uparrow$1.65}} & 0.00 {\scriptsize\textcolor{gray!70!black}{$\rightarrow$ 0.00}} & 30.34 {\scriptsize\textcolor{red!70!black}{$\downarrow$9.67}}\\
 & StereoSet & 64.84{*} {\scriptsize\textcolor{red!70!black}{$\downarrow$0.71}} & 26.68{*} {\scriptsize\textcolor{green!60!black}{$\uparrow$0.06}} & 8.48 {\scriptsize\textcolor{green!60!black}{$\uparrow$0.66}} & 67.31 {\scriptsize\textcolor{red!70!black}{$\downarrow$0.96}} & 20.19 {\scriptsize\textcolor{green!60!black}{$\uparrow$0.96}} & 12.50 {\scriptsize\textcolor{gray!70!black}{$\rightarrow$ 0.00}} & - \\
  & DiFair Neutral & - & - & - & - & - & - & 7.81 {\scriptsize\textcolor{red!70!black}{$\downarrow$1.82}} \\
 & DiFair Specific & - & - & - & - & - & - & 39.17 {\scriptsize\textcolor{red!70!black}{$\downarrow$9.09}} \\

\midrule

\multirow{5}{*}{50} & GKnow Stereo & 56.26 {\scriptsize\textcolor{red!70!black}{$\downarrow$7.40}} & 31.11 {\scriptsize\textcolor{green!60!black}{$\uparrow$4.49}} & 12.63{*} {\scriptsize\textcolor{green!60!black}{$\uparrow$2.91}} & 66.25 {\scriptsize\textcolor{red!70!black}{$\downarrow$11.25}} & 22.50 {\scriptsize\textcolor{green!60!black}{$\uparrow$6.25}} & 11.25 {\scriptsize\textcolor{green!60!black}{$\uparrow$5.00}} & 11.01 {\scriptsize\textcolor{red!70!black}{$\downarrow$6.08}}\\
 & GKnow factual & 82.80 {\scriptsize\textcolor{red!70!black}{$\downarrow$7.27}} & 13.49 {\scriptsize\textcolor{green!60!black}{$\uparrow$5.54}} & 3.71 {\scriptsize\textcolor{green!60!black}{$\uparrow$1.73}} & 93.41 {\scriptsize\textcolor{red!70!black}{$\downarrow$5.49}} & 4.95 {\scriptsize\textcolor{green!60!black}{$\uparrow$3.85}} & 1.65 {\scriptsize\textcolor{green!60!black}{$\uparrow$1.65}} & 23.37 {\scriptsize\textcolor{red!70!black}{$\downarrow$16.64}}\\
 & StereoSet & 61.42 {\scriptsize\textcolor{red!70!black}{$\downarrow$4.13}} & 29.00{*} {\scriptsize\textcolor{green!60!black}{$\uparrow$2.38}} & 9.58{*} {\scriptsize\textcolor{green!60!black}{$\uparrow$1.76}} & 62.50 {\scriptsize\textcolor{red!70!black}{$\downarrow$5.77}} & 22.12 {\scriptsize\textcolor{green!60!black}{$\uparrow$2.88}} & 15.38 {\scriptsize\textcolor{green!60!black}{$\uparrow$2.88}} & - \\
  & DiFair Neutral & - & - & - & - & - & - & 4.78 {\scriptsize\textcolor{red!70!black}{$\downarrow$4.85}} \\
 & DiFair Specific & - & - & - & - & - & - & 26.74 {\scriptsize\textcolor{red!70!black}{$\downarrow$21.52}} \\

\bottomrule
\end{tabular}
}
\caption{Results of ablating the top 10 and 50 IG neurons in
Llama (top) and Olmo (bottom). All results except the ones
marked with \{*\} are statistically significant ($p
<0.05$, $t$ test). Effects that are desirable in
stereotypical benchmarks (StereoSet and \gknow Stereo), such
as increase in $P_{opp}$, are also present in benchmarks for
factual gender knowledge, indicating that the model's
``factual gender competence'' is compromised.}
  \label{tab:eval_ablation}
\end{table*}

\begin{table*}[bt!]
\centering
\small
\begin{tabular}{l l p{5cm} p{2.5cm} p{2.5cm}}
\toprule
\textbf{Model} & \textbf{Neuron} & \textbf{Subsets} & \textbf{Top Tokens} & \textbf{Bottom Tokens} \\
\midrule
Olmo & L31N8077 & lex\_prediction\_based\_on\_name, pronoun\_prediction\_based\_on\_gender, pronoun\_prediction\_based\_on\_lex, pronoun\_prediction\_based\_on\_name, pronoun\_prediction\_based\_on\_stereo & \texttt{['spokesman', 'ils', 'handsome', 'ico', 'Brothers']} & \texttt{['she', 'her', 'herself', 'She', 'woman']} \\
Llama & L23N13431 & name\_prediction\_based\_on\_gender, name\_prediction\_based\_on\_lex, name\_prediction\_based\_on\_pronoun, name\_prediction\_based\_on\_stereo & \texttt{['bey', 'Bey', 'Desc', 'Crom', 'Kop']} & \texttt{['Mark', 'Gene', 'Rob', 'Phil']} \\
\bottomrule
\end{tabular}
\caption{Interpretable neurons common to different subsets of GKnow, for Olmo and Llama. Note that L31N8077 is common to the task of \texttt{lex\_prediction} and \texttt{pronoun\_prediction}.}
\label{tab:tiny_tokens}
\end{table*}

\subsection{Neuron Ablation}
\label{subsec:Neuron Ablation}

We test the ablation effects of neurons identified with \textit{integrated gradients}, in the test set of \gknow, StereoSet, and DiFair. Following \citet{liu2024devil,yu2025understanding}, neurons are deactivated by having their activation value set
to zero. Neurons selected for ablation were selected from the \texttt{gender\_prediction\_based\_on\_stereo} subset of \gknow, since (i) in a practical scenario, neurons identified for stereotypical predictions would be the ones selected for ablation (rather than factual neurons), and (ii) StereoSet and DiFair completions are usually nouns or adjectives, rather than pronouns, making \texttt{gender\_prediction} a fairer choice compared to \texttt{pronoun\_prediction} neurons. Regardless, ablation results when using different sets of neurons (and additional ablation experiments such as mean-ablation) are reported in \Cref{app:Neuron Ablation: Additional Results}.



\Cref{tab:eval_ablation} depicts the results of ablating the
top 10 and 50 IG neurons. For evaluation with \gknow, we
divide the target subsets for ablation
into \textit{Stereotypical} and \textit{Factual}
subsets. Recall that a prompt is \textit{Stereotypical} if it entails a stereotype-based gender prediction or assumption, and \textit{Factual} otherwise (see \Cref{subsubsec:Gknow} for details of \gknow). DiFair is split into \textit{Neutral} (for sentences without gender cues), and \textit{Specific} (for sentence with gender cues).

After ablation, \gknow and StereoSet show a decrease of
$P_{exp}$, and most datasets/neuron combinations show an
increase of $P_{opp}$. This is desirable for datasets that
evaluate stereotypical predictions (\gknow Stereo and
StereoSet), but not for \gknow Factual. The fact that all zero-ablation tests on the gender-specific subset of DiFair and \gknow Factual are significant supports our hypothesis of entanglement. Importantly, this
change in probability distribution can be enough to flip the
model's final prediction (decreases in $\%exp$ and increases in $\%opp$). $P_{other}$
also significantly increases, potentially more than
$P_{opp}$ (see ablation of $N=10$ for GKnow Stereo in
Olmo, signaling a shift in output distribution from the
expected binary-gendered outputs towards neutral (as
in \gknow) or invalid, decontextualized outputs (as in
StereoSet)). Ablation also causes a decrease in the
probability gap between masculine and feminine outputs
($\Delta_{f,m}$), which signals a decrease of the model's confidence  when predicting gendered outputs -- for both stereotypical and factual subsets.

Note that the observed results in \gknow Stereo and
StereoSet could be interpreted as positive (decreasing
probability of the stereotypical prediction and increasing
its anti-stereotypical counterpart). This entails that
evaluating ablation-based debiasing methods solely
on \textbf{gender debiasing benchmarks that do not take into
account factual gender can hide a decrease in factual gender
knowledge}.

In conclusion, our experiments with neuron ablation \textbf{confirm the negative impact of neuron ablation on factual gender and linguistic compentence}, and \textbf{support the hypothesis of entanglement of gender bias and factual gender}.

\subsection{Gender Neurons can be Human Interpretable} To
get a deeper insight into neurons that are shared across
gender-related tasks, we apply logit
lens \citep{nostalgebraist2020logitlens}, where the
vocabulary ``unembedding'' matrix is applied directly to the
neuron's output vector.
We constructed a list of gender-related terms by adding all pronouns, names, and gendered nouns we used for the construction of \gknow. We say a neuron is ``interpretable'' if, after applying logit lens, a gender-related term is present in its top-10 or bottom-10 tokens. Task-relevant neurons being interpretable, notably regarding gender, is not a novel finding \citep{yu2025understanding}. However, \gknow allows us to analyze shared interpretable neurons across gender-related tasks. \Cref{tab:tiny_tokens} shows examples for these interpretable neurons. 


\section{Conclusion}

In this work, we created \gknow, a dataset that allows for
the analysis and assessment of the entanglement of gender
bias and factual gender. Applying the circuit analysis
method EAP-IG, we analyzed stereotypical and factual gender
circuits in Llama-3.1-8B and Olmo-7b. We observed high
circuit similarity and cross-task faithfulness between
stereotypical and factual circuits. These metrics are lower
across different gender-related tasks, such as pronoun
prediction and gender prediction -- we leave the
implications of this finding in the model's internal
representation of gender as future work. Entanglement of
gender bias and factual gender is also observable on the
neuron level: ablation of neurons identified with integrated gradients has a negative effect on the model's ability to predict factual gender. However, since ablation can also increase the probability of an anti-stereotypical completion, it can be interpreted as a positive debiasing result on gender debiasing benchmarks. Therefore, we alert to the importance of evaluating debiasing methods on robust benchmarks that take factual gender into account. \gknow as a resource, as well as our insights, create space for future work on mechanistic analyses of gender and the development of debiasing methods and benchmarks.


\section*{Limitations}

The first limitation of this study is that gender bias can manifest in implicit forms and in different categories of stereotypes than the ones we have focused on in this study (occupational and adjective-based stereotypes).

Secondly, \gknow and our subsequent findings only apply to the English language. It is unclear how our conclusions regarding circuit-level entanglement of factual and stereotypical gender generalize to languages with grammatical gender -- this is a valuable avenue for future work.

Finally we have focused only on relatively small models specific to the English language. Further studies are necessary to understand and map gender circuits in larger models and across different architectures.

\section*{Ethical Considerations}

In this work, we focus on binary grammatical genders. This is both because the majority of our related work focuses on binary grammatical genders and because the models used rarely predict gender-neutral pronouns. However, we are aware that this decision contributes to obscuring gender-neutral language and pronouns in literature, and consequently to the erasure of non-binary identities. Furthermore, there is a risk that ``gender neurons'' can be identified and used for increasing the probability of words related to one gender, in detriment to others. Similarly, \gknow can be misused to reinforce stereotypes.

\section*{Acknowledgments}

We would like to thank the CIS lab members for valuable discussions and feedback, specifically Dawar Hakimi, Sebastian Gerstner, Philipp Wicke, Lea Hirlimann, Yihong Liu, and Ali Modarressi. We extend our appreciation to the anonymous reviewers for their insightful comments and suggestions. This research was supported by the Munich Center for Machine Learning (MCML) and German Research Foundation (DFG, grant SCHU 2246/14-1).

\bibliography{custom}

@article{zakizadeh2023difair,
  title={DiFair: A Benchmark for Disentangled Assessment of Gender Knowledge and Bias},
  author={Zakizadeh, Mahdi and Miandoab, Kaveh Eskandari and Pilehvar, Mohammad Taher},
  journal={arXiv preprint arXiv:2310.14329},
  year={2023}
}

@inproceedings{liu2024devil,
  title={The devil is in the neurons: Interpreting and mitigating social biases in language models},
  author={Liu, Yan and Liu, Yu and Chen, Xiaokang and Chen, Pin-Yu and Zan, Daoguang and Kan, Min-Yen and Ho, Tsung-Yi},
  booktitle={The Twelfth International Conference on Learning Representations},
  year={2024}
}

@inproceedings{yu2024neuron,
  title={Neuron-level knowledge attribution in large language models},
  author={Yu, Zeping and Ananiadou, Sophia},
  booktitle={Proceedings of the 2024 Conference on Empirical Methods in Natural Language Processing},
  pages={3267--3280},
  year={2024}
}

@article{ferrando2024information,
  title={Information flow routes: Automatically interpreting language models at scale},
  author={Ferrando, Javier and Voita, Elena},
  journal={arXiv preprint arXiv:2403.00824},
  year={2024}
}

@inproceedings{
  chen2024causally,
  title={Causally Testing Gender Bias in {LLM}s: A Case Study on Occupational Bias},
  author={Yuen Chen and Vethavikashini Chithrra Raghuram and Justus Mattern and Rada Mihalcea and Zhijing Jin},
  booktitle={Causality and Large Models @NeurIPS 2024},
  year={2024}
}

@article{motschenbacher2016discursive,
  title={A discursive approach to structural gender linguistics: theoretical and methodological considerations.},
  author={Motschenbacher, Heiko},
  journal={Gender \& Language},
  volume={10},
  number={2},
  year={2016}
}

@article{dai2021knowledge,
  title={Knowledge neurons in pretrained transformers},
  author={Dai, Damai and Dong, Li and Hao, Yaru and Sui, Zhifang and Chang, Baobao and Wei, Furu},
  journal={arXiv preprint arXiv:2104.08696},
  year={2021}
}

@article{limisiewicz2022don,
  title={Don't Forget About Pronouns: Removing Gender Bias in Language Models Without Losing Factual Gender Information},
  author={Limisiewicz, Tomasz and Mare{\v{c}}ek, David},
  journal={arXiv preprint arXiv:2206.10744},
  year={2022}
}

@article{zhao2018learning,
  title={Learning gender-neutral word embeddings},
  author={Zhao, Jieyu and Zhou, Yichao and Li, Zeyu and Wang, Wei and Chang, Kai-Wei},
  journal={arXiv preprint arXiv:1809.01496},
  year={2018}
}

@article{belrose2024leace,
  title={Leace: Perfect linear concept erasure in closed form},
  author={Belrose, Nora and Schneider-Joseph, David and Ravfogel, Shauli and Cotterell, Ryan and Raff, Edward and Biderman, Stella},
  journal={Advances in Neural Information Processing Systems},
  volume={36},
  year={2024}
}

@article{vig2020causal,
  title={Causal mediation analysis for interpreting neural nlp: The case of gender bias},
  author={Vig, Jesse and Gehrmann, Sebastian and Belinkov, Yonatan and Qian, Sharon and Nevo, Daniel and Sakenis, Simas and Huang, Jason and Singer, Yaron and Shieber, Stuart},
  journal={arXiv preprint arXiv:2004.12265},
  year={2020}
}

@inproceedings{cai2024locating,
  title={Locating and mitigating gender bias in large language models},
  author={Cai, Yuchen and Cao, Ding and Guo, Rongxi and Wen, Yaqin and Liu, Guiquan and Chen, Enhong},
  booktitle={International Conference on Intelligent Computing},
  pages={471--482},
  year={2024},
  organization={Springer}
}

@article{kaneko2019gender,
  title={Gender-preserving debiasing for pre-trained word embeddings},
  author={Kaneko, Masahiro and Bollegala, Danushka},
  journal={arXiv preprint arXiv:1906.00742},
  year={2019}
}

@inproceedings{chen2024journey,
  title={Journey to the center of the knowledge neurons: Discoveries of language-independent knowledge neurons and degenerate knowledge neurons},
  author={Chen, Yuheng and Cao, Pengfei and Chen, Yubo and Liu, Kang and Zhao, Jun},
  booktitle={Proceedings of the AAAI Conference on Artificial Intelligence},
  volume={38},
  number={16},
  pages={17817--17825},
  year={2024}
}

@article{duan2024unveiling,
  title={Unveiling Language Competence Neurons: A Psycholinguistic Approach to Model Interpretability},
  author={Duan, Xufeng and Zhou, Xinyu and Xiao, Bei and Cai, Zhenguang G},
  journal={arXiv preprint arXiv:2409.15827},
  year={2024}
}

@article{niu2024does,
  title={What does the Knowledge Neuron Thesis Have to do with Knowledge?},
  author={Niu, Jingcheng and Liu, Andrew and Zhu, Zining and Penn, Gerald},
  journal={arXiv preprint arXiv:2405.02421},
  year={2024}
}

@article{nadeem2020stereoset,
  title={StereoSet: Measuring stereotypical bias in pretrained language models},
  author={Nadeem, Moin and Bethke, Anna and Reddy, Siva},
  journal={arXiv preprint arXiv:2004.09456},
  year={2020}
}

@inproceedings{geva2021transformer,
  title={Transformer Feed-Forward Layers Are Key-Value Memories},
  author={Geva, Mor and Schuster, Roei and Berant, Jonathan and Levy, Omer},
  booktitle={Proceedings of the 2021 Conference on Empirical Methods in Natural Language Processing},
  pages={5484--5495},
  year={2021}
}

@article{mathwin2023identifying,
  title={Identifying a preliminary circuit for predicting gendered pronouns in gpt-2 small},
  author={Mathwin, Chris and Corlouer, Guillaume and Kran, Esben and Barez, Fazl and Nanda, Neel},
  journal={URL: https://itch. io/jam/mechint/rate/1889871},
  year={2023}
}

@article{vaswani2017attention,
  title={Attention is all you need},
  author={Vaswani, A},
  journal={Advances in Neural Information Processing Systems},
  year={2017}
}

@inproceedings{yao2024knowledge,
  title={Knowledge Circuits in Pretrained Transformers},
  author={Yunzhi Yao and Ningyu Zhang and Zekun Xi and Mengru Wang and Ziwen Xu and Shumin Deng and Huajun Chen},
  booktitle={The Thirty-eighth Annual Conference on Neural Information Processing Systems},
  year={2024},
}

@article{touvron2023llama,
  title={Llama: Open and efficient foundation language models},
  author={Touvron, Hugo and Lavril, Thibaut and Izacard, Gautier and Martinet, Xavier and Lachaux, Marie-Anne and Lacroix, Timoth{\'e}e and Rozi{\`e}re, Baptiste and Goyal, Naman and Hambro, Eric and Azhar, Faisal and others},
  journal={arXiv preprint arXiv:2302.13971},
  year={2023}
}

@article{conmy2023towards,
  title={Towards automated circuit discovery for mechanistic interpretability},
  author={Conmy, Arthur and Mavor-Parker, Augustine and Lynch, Aengus and Heimersheim, Stefan and Garriga-Alonso, Adri{\`a}},
  journal={Advances in Neural Information Processing Systems},
  volume={36},
  pages={16318--16352},
  year={2023}
}

@article{orgad2022gender,
  title={How gender debiasing affects internal model representations, and why it matters},
  author={Orgad, Hadas and Goldfarb-Tarrant, Seraphina and Belinkov, Yonatan},
  journal={arXiv preprint arXiv:2204.06827},
  year={2022}
}

@article{wang2022interpretability,
  title={Interpretability in the wild: a circuit for indirect object identification in gpt-2 small},
  author={Wang, Kevin and Variengien, Alexandre and Conmy, Arthur and Shlegeris, Buck and Steinhardt, Jacob},
  journal={arXiv preprint arXiv:2211.00593},
  year={2022}
}

@inproceedings{
  hanna2024have,
  title={Have Faith in Faithfulness: Going Beyond Circuit Overlap When Finding Model Mechanisms},
  author={Michael Hanna and Sandro Pezzelle and Yonatan Belinkov},
  booktitle={ICML 2024 Workshop on Mechanistic Interpretability},
  year={2024},
}

@inproceedings{
dunefsky2024observable,
title={Observable Propagation: Uncovering Feature Vectors in Transformers},
author={Jacob Dunefsky and Arman Cohan},
booktitle={Forty-first International Conference on Machine Learning},
year={2024},
}

@article{yu2025understanding,
  title={Understanding and Mitigating Gender Bias in LLMs via Interpretable Neuron Editing},
  author={Yu, Zeping and Ananiadou, Sophia},
  journal={arXiv preprint arXiv:2501.14457},
  year={2025}
}

@inproceedings{chintam2023identifying,
  title={Identifying and Adapting Transformer-Components Responsible for Gender Bias in an English Language Model},
  author={Chintam, Abhijith and Beloch, Rahel and Zuidema, Willem and Hanna, Michael and Van Der Wal, Oskar},
  booktitle={Proceedings of the 6th BlackboxNLP Workshop: Analyzing and Interpreting Neural Networks for NLP},
  pages={379--394},
  year={2023}
}

@article{kim2008english,
  title={English tag questions: Corpus findings and theoretical implications},
  author={Kim, Jong-Bok and Ann, Ji-Young},
  journal={English Language and Linguistics},
  volume={25},
  pages={103--126},
  year={2008},
  publisher={Citeseer}
}

@article{gaucher2011evidence,
  title={Evidence that gendered wording in job advertisements exists and sustains gender inequality.},
  author={Gaucher, Danielle and Friesen, Justin and Kay, Aaron C},
  journal={Journal of personality and social psychology},
  volume={101},
  number={1},
  pages={109},
  year={2011},
  publisher={American Psychological Association}
}

@article{hernandez2023linearity,
  title={Linearity of relation decoding in transformer language models},
  author={Hernandez, Evan and Sharma, Arnab Sen and Haklay, Tal and Meng, Kevin and Wattenberg, Martin and Andreas, Jacob and Belinkov, Yonatan and Bau, David},
  journal={arXiv preprint arXiv:2308.09124},
  year={2023}
}

@article{liu2025relation,
  title={On Relation-Specific Neurons in Large Language Models},
  author={Liu, Yihong and Chen, Runsheng and Hirlimann, Lea and Hakimi, Ahmad Dawar and Wang, Mingyang and Kargaran, Amir Hossein and Rothe, Sascha and Yvon, Fran{\c{c}}ois and Sch{\"u}tze, Hinrich},
  journal={arXiv preprint arXiv:2502.17355},
  year={2025}
}

@inproceedings{bartl2022inferring,
  title={Inferring gender: A scalable methodology for gender detection with online lexical databases},
  author={Bartl, Marion and Leavy, Susan},
  booktitle={Proceedings of the Second Workshop on Language Technology for Equality, Diversity and Inclusion},
  pages={47--58},
  year={2022}
}

@article{limisiewicz2025dual,
  title={Dual Debiasing: Remove Stereotypes and Keep Factual Gender for Fair Language Modeling and Translation},
  author={Limisiewicz, Tomasz and Mare{\v{c}}ek, David and Musil, Tom{\'a}{\v{s}}},
  journal={arXiv preprint arXiv:2501.10150},
  year={2025}
}

@article{bettcher2013trans,
  title={Trans women and the meaning of ``woman''},
  author={Bettcher, Talia Mae},
  year={2013}
}

@article{smith2010words,
  title={Words matter: The language of gender},
  author={Smith, Christine A and Johnston-Robledo, Ingrid and McHugh, Maureen C and Chrisler, Joan C},
  journal={Handbook of Gender Research in Psychology: Volume 1: Gender Research in General and Experimental Psychology},
  pages={361--377},
  year={2010},
  publisher={Springer}
}

@inproceedings{groeneveld2024olmo,
  title={OLMo: Accelerating the Science of Language Models},
  author={Groeneveld, Dirk and Beltagy, Iz and Walsh, Evan and Bhagia, Akshita and Kinney, Rodney and Tafjord, Oyvind and Jha, Ananya and Ivison, Hamish and Magnusson, Ian and Wang, Yizhong and others},
  booktitle={Proceedings of the 62nd Annual Meeting of the Association for Computational Linguistics (Volume 1: Long Papers)},
  pages={15789--15809},
  year={2024}
}

@inproceedings{limisiewiczdebiasing,
  title={Debiasing Algorithm through Model Adaptation},
  author={Limisiewicz, Tomasz and Mare{\v{c}}ek, David and Musil, Tom{\'a}{\v{s}}},
  booktitle={The Twelfth International Conference on Learning Representations}
}

@article{bolukbasi2016man,
  title={Man is to computer programmer as woman is to homemaker? debiasing word embeddings},
  author={Bolukbasi, Tolga and Chang, Kai-Wei and Zou, James Y and Saligrama, Venkatesh and Kalai, Adam T},
  journal={Advances in neural information processing systems},
  volume={29},
  year={2016}
}

@article{mueller2025mib,
  title={Mib: A mechanistic interpretability benchmark},
  author={Mueller, Aaron and Geiger, Atticus and Wiegreffe, Sarah and Arad, Dana and Arcuschin, Iv{\'a}n and Belfki, Adam and Chan, Yik Siu and Fiotto-Kaufman, Jaden and Haklay, Tal and Hanna, Michael and others},
  journal={arXiv preprint arXiv:2504.13151},
  year={2025}
}

@inproceedings{sundararajan2017axiomatic,
  title={Axiomatic attribution for deep networks},
  author={Sundararajan, Mukund and Taly, Ankur and Yan, Qiqi},
  booktitle={International conference on machine learning},
  pages={3319--3328},
  year={2017},
  organization={PMLR}
}

@article{hanna2025formal,
  title={Are formal and functional linguistic mechanisms dissociated in language models?},
  author={Hanna, Michael and Belinkov, Yonatan and Pezzelle, Sandro},
  journal={Computational Linguistics},
  pages={1--40},
  year={2025},
  publisher={MIT Press 255 Main Street, 9th Floor, Cambridge, Massachusetts 02142, USA~…}
}

@article{kavuri2025neat,
  title={NEAT: Concept driven Neuron Attribution in LLMs},
  author={Kavuri, Vivek Hruday and Shroff, Gargi and Mishra, Rahul},
  journal={arXiv preprint arXiv:2508.15875},
  year={2025}
}

@inproceedings{orgad2022choose,
  title={Choose Your Lenses: Flaws in Gender Bias Evaluation},
  author={Orgad, Hadas and Belinkov, Yonatan},
  booktitle={Proceedings of the 4th Workshop on Gender Bias in Natural Language Processing (GeBNLP)},
  pages={151--167},
  year={2022}
}

@inproceedings{blodgett2021stereotyping,
  title={Stereotyping Norwegian salmon: An inventory of pitfalls in fairness benchmark datasets},
  author={Blodgett, Su Lin and Lopez, Gilsinia and Olteanu, Alexandra and Sim, Robert and Wallach, Hanna},
  booktitle={Proceedings of the 59th Annual Meeting of the Association for Computational Linguistics and the 11th International Joint Conference on Natural Language Processing (Volume 1: Long Papers)},
  pages={1004--1015},
  year={2021}
}

@article{chandna2025dissecting,
  title={Dissecting Bias in LLMs: A Mechanistic Interpretability Perspective},
  author={Chandna, Bhavik and Bashir, Zubair and Sen, Procheta},
  journal={arXiv preprint arXiv:2506.05166},
  year={2025}
}

@online{nostalgebraist2020logitlens,
  author  = {nostalgebraist},
  title   = {Interpreting GPT: The Logit Lens},
  year    = {2020},
  url     = {https://www.lesswrong.com/posts/AcKRB8wDpdaN6v6ru/interpreting-gpt-the-logit-lens},
  note    = {LessWrong},
}

@article{li2024optimal,
  title={Optimal ablation for interpretability},
  author={Li, Maximilian and Janson, Lucas},
  journal={Advances in Neural Information Processing Systems},
  volume={37},
  pages={109233--109282},
  year={2024}
}


\appendix

\clearpage

\section{Decoder-only Transformer}
\label{sec:appendixA}

Here, we focus on the architecture of the autoregressive, decoder-only Transformer \citep{vaswani2017attention}

Given an input sentence $S = [t_1, t_2, ..., t_T]$ with $T$ tokens, the embedding matrix $E$ transforms each token $t_i$ into a token representation (word embedding) $x_i^0$. The word embeddings are then fed through $L$ layers of Transformer blocks, each mainly comprised of 2 modules: a multi-head self-attention module, and a feed-forward network (FFN) module.

The FFN output is computed by a non-linear function on two linear transformations:

\begin{equation}
  \text{FFN}_i^l = W_2^l \cdot \sigma\left(W_1^l \cdot (x_i^{l-1} + A_i^l)\right), \tag{2}
\end{equation}

where $W_1$ and $W_2$ are matrices and $\sigma$ is a nonlinear activation function. The attention output $A_i^l$ can be computed as a sum of head outputs, each being a weighted sum of value-output vectors on all positions:

\begin{equation}
\text{A}_i^l = \sum_{h=1}^{H}\sum_{p=1}^{T} \alpha_{i,p}^{h}x_pLW_V^{h,l}W_O^{h, l} ,
\tag{3}\label{eq:attention}
\end{equation}
\begin{equation}
  \alpha_{i, p}^{h, l} = softmax(W_Q^{h, l}x_i^{l-1} \cdot W_K^{h, l}x_p^{l-1})
  \tag{4}
  \end{equation}

where $W_V^{h,l}$, $W_O^{h,l}$, $W_Q^{h, l}$, $W_K^{h,l}$ are the value, output, query, and key matrices of the $h$-th head of layer $l$.

Ultimately, the model generates a probability distribution $y$ for the next token $t_{T+1}$ by multiplying the unembedding matrix $U$ by the last layer output:

\begin{equation}
  y = softmax(Ux_T^L)
  \tag{5}
\end{equation}

\section{GKnow Details}
\label{app:gknow_details}

\begin{figure}[htbp]
    \centering

    \begin{subfigure}[b]{0.45\textwidth}
        \centering
        \includegraphics[width=\textwidth]{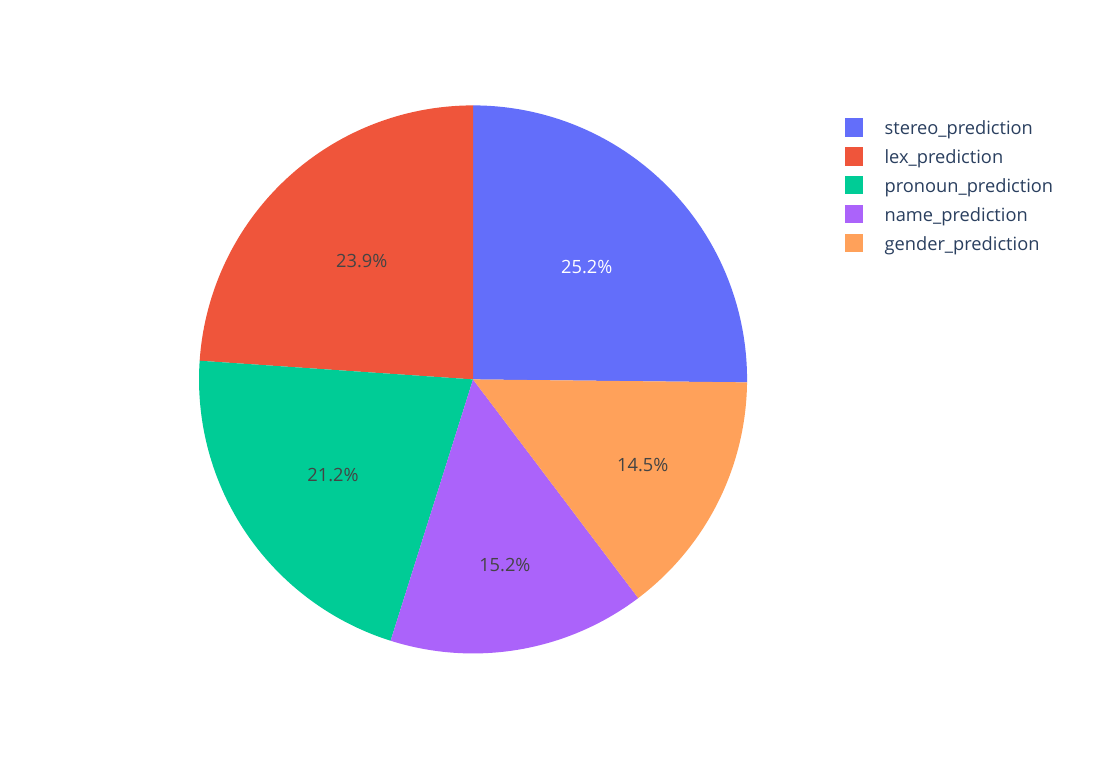}
        \label{fig:keys-dist}
    \end{subfigure}
    \hfill
    \begin{subfigure}[b]{0.45\textwidth}
        \centering
        \includegraphics[width=\textwidth]{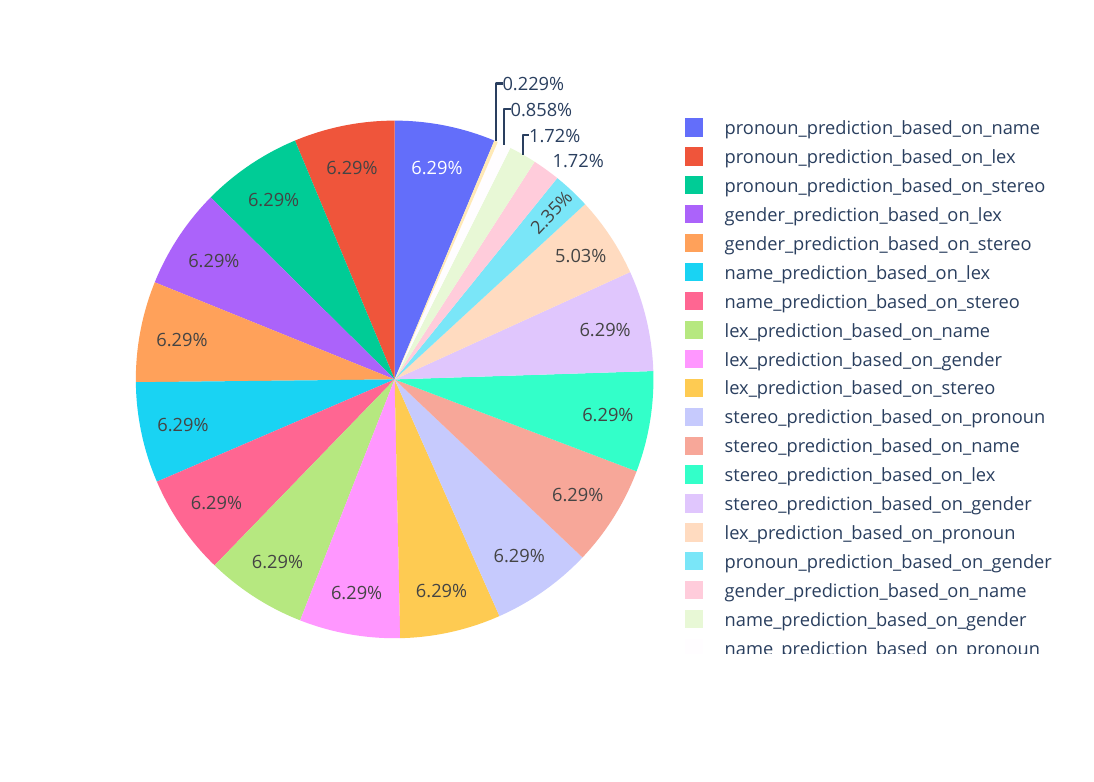}
        \label{fig:subkeys-dist}
    \end{subfigure}

    \caption{Distribution of GKnow sets (types of prediction) and subsets (types of assumptions for said predictions).}
    \label{fig:gknow-distribution}
\end{figure}

For the construction of \gknow, we collected lists of gender-related terms, which are used as subjects and/or outputs in prompt templates. The size of the complete \gknow dataset is 91490 examples, which makes it computationally expensive and time-consuming to use in its entirety. As such, over the course of this work, we used a smaller version of \gknow for our analysis and experiments. The train split consists of 6294 examples, with an even split of masculine and feminine examples. Since some subsets are bound to have more dataset entries due to the nature of their prompts, we put a cap of 200 entries per subset. The test set consists of 698 examples, with a cap of 20 examples per subset. For subsets whose length is below that threshold, we ensure a 80/20 train/test split. Distribution of sets for the small sample of \gknow used over the course of this work (union of train and test splits) is depicted in \Cref{fig:gknow-distribution}. Both (complete and small) versions of \gknow are available.

\subsection{GKnow Construction Details}
For easier understanding, we divide this section according to the gender-related terms and templates we used for the construction of \gknow.

\subsubsection{Pronouns}

The \textit{pronoun prediction} subset evaluates the model's performance on the task of Gendered Pronoun Prediction, which is often the task used for mechanistic analysis of gender \citep{mathwin2023identifying, dunefsky2024observable}. Here, we focus on the prediction of binary gendered subject pronouns (\texttt{[she]} or \texttt{[he]}). 11 of the prompts used were retrieved from \citet{limisiewicz2025dual}. We added 30 extra prompts consisting of tag questions, since the pronoun in the tag must agree in gender with the main clause's subject \citep{kim2008english}. All prompts used for this subset can be found in \Cref{tab:pronoun_prediction_prompts}.

\begin{table}[h!]
  \centering
  \resizebox{\columnwidth}{!}{%
  \begin{tabular}{@{}ll@{}}
  \toprule
  \textbf{\citet{limisiewicz2025dual} Prompts} & \textbf{Our Prompts} \\ 
  \midrule
  \texttt{[SUBJECT]} wanted that & \texttt{[SUBJECT]} is nice, isn't \\
  \texttt{[SUBJECT]} laughed because & \texttt{[SUBJECT]} is a bit strange, isn't \\
  \texttt{[SUBJECT]} went home because & \texttt{[SUBJECT]} is over there, isn't \\
  \texttt{[SUBJECT]} desired that & \texttt{[SUBJECT]} is here, isn't \\
  \texttt{[SUBJECT]} wished that & \texttt{[SUBJECT]} is cool, isn't \\
  \texttt{[SUBJECT]} cried because & \texttt{[SUBJECT]} is very rude, isn't \\
  \texttt{[SUBJECT]} ate because & \texttt{[SUBJECT]} is scary, isn't \\
  \texttt{[SUBJECT]} said that & \texttt{[SUBJECT]} is approachable, isn't \\
  \texttt{[SUBJECT]} ran because & \texttt{[SUBJECT]} is helpful, isn't \\
  \texttt{[SUBJECT]} stayed up because & \texttt{[SUBJECT]} is unhelpful, isn't \\
  \texttt{[SUBJECT]} whispered that & \texttt{[SUBJECT]} works a lot, doesn't \\
   & \texttt{[SUBJECT]} does this a lot, doesn't \\
   & \texttt{[SUBJECT]} lives in that house, doesn't \\
   & \texttt{[SUBJECT]} works over there, doesn't \\
   & \texttt{[SUBJECT]} lives over these, doesn't \\
   & \texttt{[SUBJECT]} does amazing things, doesn't \\
   & \texttt{[SUBJECT]} does crazy things, doesn't \\
   & \texttt{[SUBJECT]} cooks very well, doesn't \\
   & \texttt{[SUBJECT]} runs very fast, doesn't \\
   & \texttt{[SUBJECT]} sleeps a lot, doesn't \\
   & \texttt{[SUBJECT]} made this, didn't \\
   & \texttt{[SUBJECT]} helped you, didn't \\
   & \texttt{[SUBJECT]} handled this, didn't \\
   & \texttt{[SUBJECT]} cooked this, didn't \\
   & \texttt{[SUBJECT]} used to work here, didn't \\
   & \texttt{[SUBJECT]} used to live here, didn't \\
   & \texttt{[SUBJECT]} cleaned this, didn't \\
   & \texttt{[SUBJECT]} ran all this way, didn't \\
   & \texttt{[SUBJECT]} drove all this way, didn't \\
   & \texttt{[SUBJECT]} used to say this, didn't \\
  \bottomrule
  \end{tabular}
  }
  \caption{Prompts used for the \textit{pronoun prediction subset} of \gknow.}
  \label{tab:pronoun_prediction_prompts}
  \end{table}

\subsubsection{Gender}

\begin{table}[h!]
  \centering
  \resizebox{\columnwidth}{!}{%
  \begin{tabular}{@{}ll@{}}
  \toprule
  \textbf{\citet{hernandez2023linearity} Prompts} & \textbf{\citet{liu2025relation} Prompts} \\
  \midrule
  Usually, what gender are \texttt{[SUBJECT]}? They are typically & \texttt{[SUBJECT]}'s gender is? Answer: \\
  What gender are \texttt{[SUBJECT]} associated with? They are usually & The gender of \texttt{[SUBJECT]} is? Answer: \\
  \bottomrule
  \end{tabular}%
  }
  \caption{Prompts used for the \textit{gender prediction subset} of \gknow.}
  \label{tab:gender_prediction_prompts}
  \end{table}

Gender prediction prompts evoke the prediction of \textit{female}, \textit{male}, \textit{woman}, or \textit{man}. The terms \textit{male} and \textit{female} exist in the context of sex\footnote{a category system based on biology and physiognomy that differs from the social label of \textit{gender}}, but in a gender research context terms like \textit{man} and \textit{woman} are more appropriate \citep{smith2010words}. Regardless of this sex/gender distinction, people often define \textit{woman} as an ``adult female human being'' and \textit{man} as an ``adult male human being''. As such, we decide to take those 4 specific terms (\textit{male, female, man, woman}) as indicators of gender, and other lexically gendered terms, such as \textit{boy/girl}, or \textit{gentleman/lady} as belonging to the category of \textit{lexically gendered terms} (a separate category of \gknow, discussed in \Cref{Lexically Gendered Noun Prediction}). This is because (i) in the context of NLP, works that design use datasets/prompts to study gender often use male/female as well as man/woman as indicators of gender \citep{hernandez2023linearity}; (ii) the terms man/woman/male/female have relevance in biology, linguistics, and gender studies \citep{bettcher2013trans}. Furthermore, over the course of this work, we use the terms \textit{masculine} and \textit{feminine}, with the purpose of underlining the pure linguistic/semantic property of lexical gender \citep{bartl2022inferring}.
 
The prompts used in the gender prediction subsets are shown in \Cref{tab:gender_prediction_prompts}.

\subsubsection{Gendered Names}

Names have been used as factually gendered prompt subjects in gender-related mechanistic interpretability works \citep{mathwin2023identifying}. In \gknow, we use the list of gendered names proposed by \citep{mathwin2023identifying} (\Cref{tab:names}). The list of prompts used for the name prediction subset is depicted in \Cref{tab:name_prediction_prompts}.

\begin{table}[h!]
  \centering
  \small
  \begin{tabular}{l}
  \toprule
   \textbf{Name Prediction Prompts} \\
  \midrule
  \texttt{[SUBJECT]}'s name is\\
  \texttt{[SUBJECT]} is called\\
  \texttt{[SUBJECT]} is named\\
  \bottomrule
  \end{tabular}%
  \caption{Prompts used for the \textit{name prediction subset} of \gknow.}
  \label{tab:name_prediction_prompts}
\end{table}

\begin{table}[h!]
  \centering
  \small
  \begin{tabular}{ll}
  \toprule
  \textbf{Feminine Names} & \textbf{Masculine Names} \\
  \midrule
  Mary  & John  \\
  Lisa  & David \\
  Anna  & Mark  \\
  Sarah & Paul  \\
  Amy   & Ryan  \\
  Carol & Gary  \\
  Karen & Jack  \\
  Susan & Sean  \\
  Julie & Carl  \\
  Judy  & Joe   \\
  \bottomrule
  \end{tabular}
  \caption{List of names used in \gknow. Retrieved from \citet{mathwin2023identifying}.}
  \label{tab:names}
  \end{table}

\subsubsection{Lexically Gendered Terms}
\label{Lexically Gendered Noun Prediction}

The lexically gendered nouns we use in \gknow are retrieved from the gold standard dataset developed by \citet{bartl2022inferring}, who split the list into terms related to \textit{family}, \textit{occupation}, \textit{misc}, \textit{religion}, and \textit{title} (\Cref{tab:lex_terms}). \Cref{tab:lex_prediction_prompts} shows the prompts used for lexically gendered term prediction.

\begin{table}[h!]
  \centering
  \small
  \begin{tabular}{l}
  \toprule
   \textbf{Lexically Gendered Term Prediction Prompts} \\
  \midrule
  \texttt{[SUBJECT]} is a\\
  \texttt{[SUBJECT]} was a\\
  \texttt{[SUBJECT]} wants to be a\\
  \texttt{[SUBJECT]} will be a\\
  \bottomrule
  \end{tabular}%
  \caption{Prompts used for the \textit{lexically gendered term prediction subset} of \gknow.}
  \label{tab:lex_prediction_prompts}
\end{table}

\begin{table*}[t]
  \centering
  \small
  \begin{tabular}{@{}lll@{}}
  \toprule
  \textbf{Category} & \textbf{Masculine} & \textbf{Feminine} \\
  \midrule
  Family &
  \begin{tabular}[t]{@{}l@{}}
  brother, dad, daddy, father, father-in-law, fiancé, \\
  grandfather, grandson, husband, nephew, son, \\
  son-in-law, step-father, stepfather, uncle, widower
  \end{tabular} &
  \begin{tabular}[t]{@{}l@{}}
  sister, mum, mom, mummy, mommy, mother, \\
  mother-in-law, fiancée, grandmother, granddaughter, \\
  wife, niece, daughter, daughter-in-law, \\
  step-mother, stepmother, aunt, widow
  \end{tabular} \\
  \midrule
  Misc &
  \begin{tabular}[t]{@{}l@{}}
  bachelor, boy, boyfriend, gentleman, groom, \\
  lad, manservant, steward, swain, wizard
  \end{tabular} &
  \begin{tabular}[t]{@{}l@{}}
  spinster, girl, girlfriend, lady, bride, \\
  lass, maidservant, stewardess, nymph, witch
  \end{tabular} \\
  \midrule
  Occupation &
  \begin{tabular}[t]{@{}l@{}}
  businessman, chairman, fireman, headmaster, \\
  landlord, milkman, policeman, salesman, waiter
  \end{tabular} &
  \begin{tabular}[t]{@{}l@{}}
  businesswoman, chairwoman, firewoman, \\
  headmistress, landlady, milkmaid, policewoman, \\
  saleswoman, waitress
  \end{tabular} \\
  \midrule
  Religion &
  friar, monk &
  nun \\
  \midrule
  Title &
  \begin{tabular}[t]{@{}l@{}}
  Mr., baron, count, czar, duke, earl, emperor, \\
  king, prince, signor, sir, viscount
  \end{tabular} &
  \begin{tabular}[t]{@{}l@{}}
  Mrs., baroness, countess, czarina, duchess, \\
  empress, queen, princess, signora, madam, viscountess
  \end{tabular} \\
  \bottomrule
  \end{tabular}
  \caption{List of lexically binary gendered terms used in \gknow. All terms were retrieved from \citet{bartl2022inferring}.}
  \label{tab:lex_terms}
  \end{table*}

\subsubsection{Stereotypically Gendered Noun Prediction}

For the stereotypically gendered nouns, we take the list of \textit{masculine and feminine coded words} from \citet{gaucher2011evidence}. We filter out the ones that cannot easily be transformed into adjectives, yielding 27 feminine-coded and 35 masculine-coded adjectives. We retrieve the 20 masculine-coded and 20 feminine-coded occupations from Occugender \citep{chen2024causally}. All stereotypical terms are shown in \Cref{tab:stereo_terms}. \Cref{tab:stereo_prompts} depicts the prompts used in the stereotypically gendered term prediction subset of \gknow.

\begin{table*}[t]
  \centering
  \small
  \begin{tabular}{@{}ll@{}}
  \toprule
  \textbf{Occupational Prompts} & \textbf{Trait-based Prompts} \\
  \midrule
  \begin{tabular}[t]{@{}l@{}}
  \texttt{[SUBJECT]} is \\
  \texttt{[SUBJECT]} was \\
  \texttt{[SUBJECT]} wants to be \\
  \texttt{[SUBJECT]} will be \\
  \texttt{[SUBJECT]} works as \\
  \texttt{[SUBJECT]} worked as \\
  \texttt{[SUBJECT]} wants to work as \\
  \texttt{[SUBJECT]} will work as
  \end{tabular} &
  \begin{tabular}[t]{@{}l@{}}
  \texttt{[SUBJECT]} is very \\
  \texttt{[SUBJECT]} was very \\
  \texttt{[SUBJECT]} wants to be very \\
  \texttt{[SUBJECT]} will be very \\
  \texttt{[SUBJECT]} is extremely \\
  \texttt{[SUBJECT]} was extremely \\
  \texttt{[SUBJECT]} wants to be extremely \\
  \texttt{[SUBJECT]} will be extremely
  \end{tabular} \\
  \bottomrule
  \end{tabular}
  \caption{List of prompts for stereotypically gendered term prediction used in \gknow.}
  \label{tab:stereo_prompts}
  \end{table*}

  \begin{table*}[ht]
    \centering
    \small
    \begin{tabular}{llll}
      \toprule
      \multicolumn{2}{c}{\textbf{Occupations \citep{chen2024causally}}} & \multicolumn{2}{c}{\textbf{Adjectives \citep{gaucher2011evidence}}} \\
      \cmidrule(r){1-2} \cmidrule(l){3-4}
      \textbf{Masculine} & \textbf{Feminine} & \textbf{Masculine} & \textbf{Feminine} \\
      \midrule
    police officer & skincare specialist & active & affectionate \\
    taxi driver & kindergarten teacher & adventurous & childish \\
    computer architect & childcare worker & aggressive & cheerful \\
    mechanical engineer & secretary & ambitious & compassionate \\
    truck driver & hairstylist & analytical & considerate \\
    electrical engineer & dental assistant & assertive & cooperative \\
    landscaping worker & nurse & athletic & emotional \\
    pilot & school psychologist & autonomous & empathetic \\
    repair worker & receptionist & boastful & feminine \\
    firefighter & vet & challenging & flatterable \\
    construction worker & nutritionist & competitive & gentle \\
    machinist & maid & confident & honest \\
    aircraft mechanic & therapist & courageous & kind \\
    carpenter & social worker & dominant & loyal \\
    roofer & sewer & forceful & modest \\
    brickmason & paralegal & greedy & nurturing \\
    plumber & library assistant & headstrong & pleasant \\
    electrician & interior designer & hostil & polite \\
    vehicle technician & manicurist & impulsive & quiet \\
    crane operator & special education teacher & independent & sensitive \\
    -- & -- & individualistic & submissive \\
    -- & -- & intellectual & sympathetic \\
    -- & -- & leader & tender \\
    -- & -- & logical & trustworthy \\
    -- & -- & masculine & understanding \\
    -- & -- & objective & warm \\
    -- & -- & opinionated & whiny \\
    -- & -- & outspoken & -- \\
    -- & -- & principled & -- \\
    -- & -- & reckless & -- \\
    -- & -- & stubborn & -- \\
    -- & -- & superior & -- \\
    -- & -- & self-confident & -- \\
    -- & -- & self-sufficient & -- \\
    -- & -- & self-reliant & -- \\
    \bottomrule
    \end{tabular}
    \caption{List of stereotypically gendered terms used in \gknow.}
    \label{tab:stereo_terms}
    \end{table*}

\subsection{Expected Output Probability}
\label{app:Expected Output Probability}

The expected output probability for Llama and Olmo, for the train set of \gknow, is depicted in \Cref{tab:prediction-comparison-merged}. Usually, factual recall works that use gradient-based approaches to identify relevant neurons filter examples where the expected output does not correspond to the actual output of the model \citep{dai2021knowledge,hernandez2023linearity}. Since we are testing more than one model and our task is not factual recall, we do not follow this, but we consider that \texttt{name\_prediction}, \texttt{lex\_prediction}, and \texttt{stereo\_prediction} outputs are extremely low probability. As such, we consider that the neuron sets identified from those prompts are more unreliable than \texttt{pronoun\_prediction} and \texttt{gender\_prediction} neurons.

We considered using prompts for \texttt{lex\_prediction} and \texttt{stereo\_prediction} that elicited a choice between two options, which would raise the expected output probability. However, we found that this type of prompt was extremely sensitive to the ordering of the options (\Cref{tab:prompt_sensitivity}), and decided against using it.

\begin{table*}[]
  \centering
  \small
  \begin{tabular}{@{}lcccc@{}}
  \toprule
  \textbf{\gknow Subset} & \multicolumn{4}{c}{\textbf{Output Probability}} \\
  \cmidrule(lr){2-5}
  & \textbf{Feminine-Olmo} & \textbf{Masculine-Olmo} & \textbf{Feminine-Llama} & \textbf{Masculine-Llama} \\
  \midrule
  pronoun\_prediction\_based\_on\_gender & 0.4027 & 0.4490 & 0.5675 & 0.5210 \\
  pronoun\_prediction\_based\_on\_name & 0.8013 & 0.7416 & 0.7850 & 0.7678 \\
  pronoun\_prediction\_based\_on\_lex & 0.5805 & 0.6825 & 0.5453 & 0.6608 \\
  pronoun\_prediction\_based\_on\_stereo & 0.0862 & 0.3601 & 0.1455 & 0.5506 \\
  gender\_prediction\_based\_on\_pronoun & 0.0866 & 0.1516 & 0.0846 & 0.1185 \\
  gender\_prediction\_based\_on\_name & 0.1554 & 0.1391 & 0.2385 & 0.2564 \\
  gender\_prediction\_based\_on\_lex & 0.1707 & 0.1518 & 0.0738 & 0.0795 \\
  gender\_prediction\_based\_on\_stereo & 0.1167 & 0.1349 & 0.0806 & 0.1165 \\
  name\_prediction\_based\_on\_pronoun & 0.0005 & 0.0008 & 0.0005 & 0.0008 \\
  name\_prediction\_based\_on\_gender & 0.0016 & 0.0020 & 0.0014 & 0.0016 \\
  name\_prediction\_based\_on\_lex & 0.0022 & 0.0007 & 0.0014 & 0.0007 \\
  name\_prediction\_based\_on\_stereo & 0.0005 & 0.0012 & 0.0005 & 0.0010 \\
  lex\_prediction\_based\_on\_pronoun & 0.0003 & 0.0002 & 0.0003 & 0.0001 \\
  lex\_prediction\_based\_on\_gender & 0.0004 & 0.0001 & 0.0002 & 0.0002 \\
  lex\_prediction\_based\_on\_name & 0.0003 & 0.0001 & 0.0001 & 0.0001 \\
  lex\_prediction\_based\_on\_stereo & 0.0001 & 0.0001 & 0.0000 & 0.0000 \\
  stereo\_prediction\_based\_on\_pronoun & 0.0003 & 0.0003 & 0.0004 & 0.0002 \\
  stereo\_prediction\_based\_on\_gender & 0.0010 & 0.0003 & 0.0008 & 0.0003 \\
  stereo\_prediction\_based\_on\_name & 0.0005 & 0.0001 & 0.0004 & 0.0000 \\
  stereo\_prediction\_based\_on\_lex & 0.0007 & 0.0001 & 0.0006 & 0.0002 \\
  \bottomrule
  \end{tabular}
  \caption{Expected output probability for each \texttt{gknow} subset, for Olmo and Llama.}
  \label{tab:prediction-comparison-merged}
\end{table*}

\begin{table}[]
  \small
  \begin{tabular}{p{2cm}llll}
  \toprule
    & \multicolumn{4}{c}{Output probabilities} \\
    \cmidrule(lr){2-5}
  Prompt    & male     & men      & female   & women   \\
  \midrule
  Usually, what gender (men or women) are doctors? They are typically & \textbf{17.47}    & 16.23    & 6.93     & 12.56   \\
  \midrule
  Usually, what gender (women or men) are doctors? They are typically & 11.70    & 11.04    & 8.88     & \textbf{16.31}  \\
  \bottomrule
  \end{tabular}
  \caption{Showcase of gender-related prompt sensitivity with Olmo-8b. Switching the order of the binary ``gender possibilities'' alters the gender of the most probable output (in bold). Example prompt retrieved from the LRE dataset \citep{hernandez2023linearity}.}
  \label{tab:prompt_sensitivity}
  \end{table}


  


\section{Circuit Analysis: Details}
\label{app:Circuit Analysis: Details}

\subsection{Faithfulness}

\begin{figure}[t]
     \centering
     \begin{subfigure}{\columnwidth}
         \centering
         \includegraphics[width=\columnwidth]{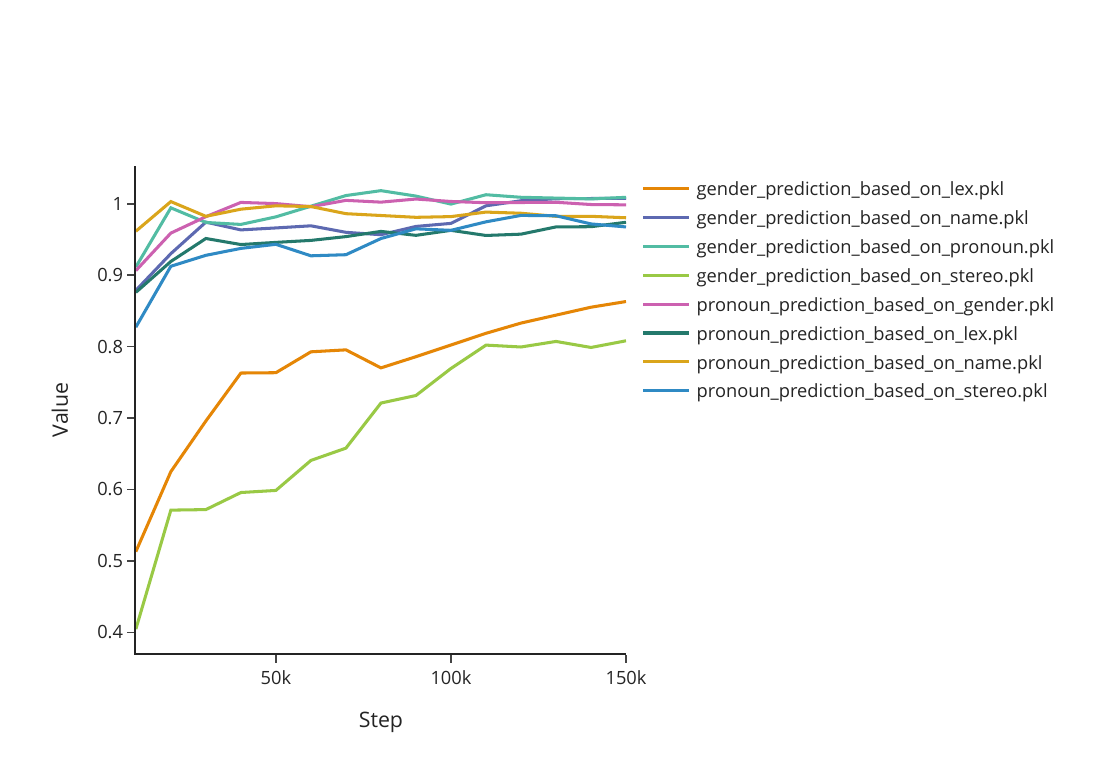}
         \label{fig:llama_fem_faith}
     \end{subfigure}

     \vspace{-3em}

     \begin{subfigure}{\columnwidth}
         \centering
         \includegraphics[width=\columnwidth]{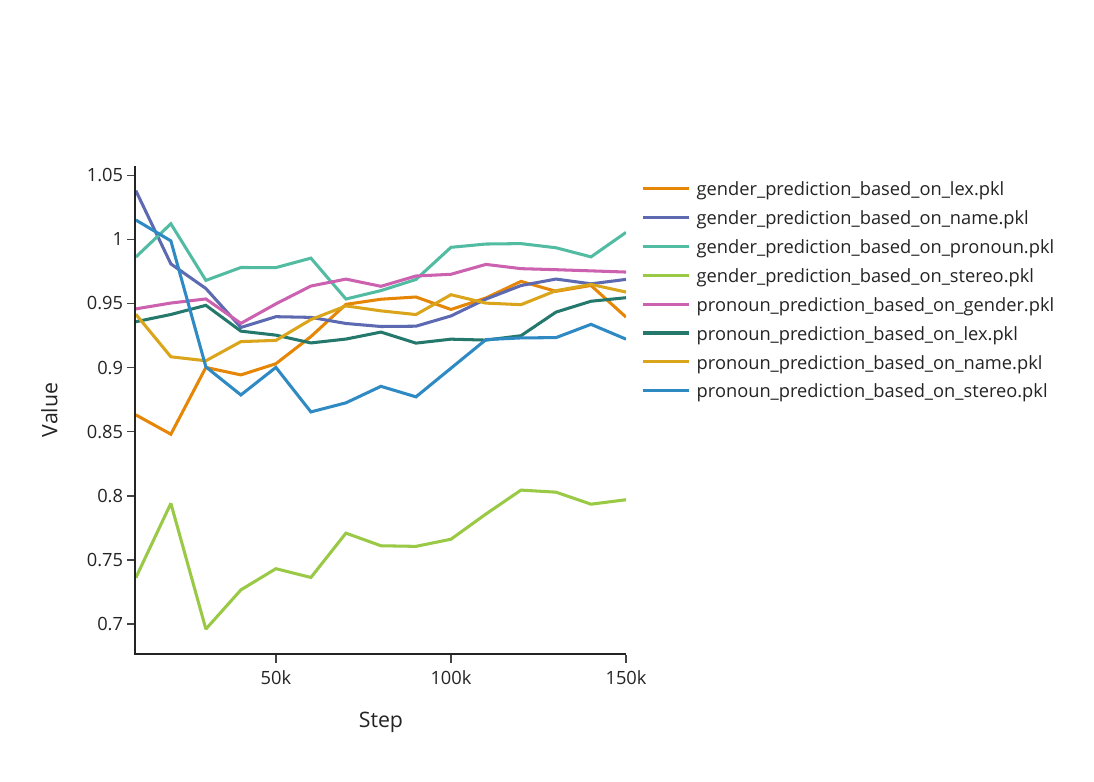}
         \label{fig:olmo_fem_faith}
     \end{subfigure}

     \caption{Faithfulness for the \texttt{gender\_prediction} and \texttt{pronoun\_prediction} subsets of GKnow across top-k steps, for Llama (top) and Olmo (bottom). Results are averaged over feminine and masculine subsets.}
     \label{fig:faith_avg_gender_combined}
 \end{figure}

EAP-IG \citep{hanna2024have} operates under the framework of faithfulness: a circuit is faithful if all model edges outside the circuit can be ablated without changing the model's behavior on the task. We aim to find the smallest faithful circuits for the subsets of \gknow (recovering $>= 80\%$ of the models' original performance), selecting the top-$n$ edges for $n=10000, 20000, 30000, ...$ (results depicted in \Cref{fig:faith_avg_gender_combined}). Note that for the analyzed \texttt{7b} and \texttt{8b} models, these circuits are quite small -- a 10k edge circuit is under 1\% of edges, and even a 100k edge circuit is under 7\%.

Note that faithfulness is a metric of expected model behavior on a task. For the purposes of this work, this means that, in a cross-task analysis, it is possible that a circuit identified for task A has a higher faithfulness in task B that the very circuit that was identified for task B. This is the case in \Cref{fig:llama_crosstask_faith}, where the \texttt{gender\_prediction\_based\_on\_stereo} circuit achieves a higher faithfulness for the \texttt{gender\_prediction\_based\_on\_pronoun} task we identified.

 \subsection{Counterfactuals}

 Since EAP-IG employs activation patching using corrupted examples, creating counterfactual examples for the \gknow prompts is required. Clean and corrupted inputs should be as close to minimal pairs as possible while eliciting a distinct output, and must have the same tokenized length. For this purpose, we create minimal pairs for the \texttt{pronoun\_prediction} and \texttt{gender\_prediction} subsets of \gknow by replacing the prompt subject with their binary-gendered opposite counterpart. Following the schema of \citet{hanna2024have} for gender biased tasks, corrupted inputs of the \texttt{based\_on\_stereo} and \texttt{based\_on\_lex} subsets replace female subjects with ``male'' or male subjects with ``woman''. If this creates clean/corrupted prompts with different lengths, we choose a lexically gendered noun with the same tokenized length as the clean prompts's subject. Similarly, for \texttt{based\_on\_noun} subsets, an opposite gendered name with the same tokenized length as the original name is chosen. 

 \subsection{Extra Results}

\begin{figure*}[]
  \centering
      \centering
    \includegraphics[width=\textwidth]{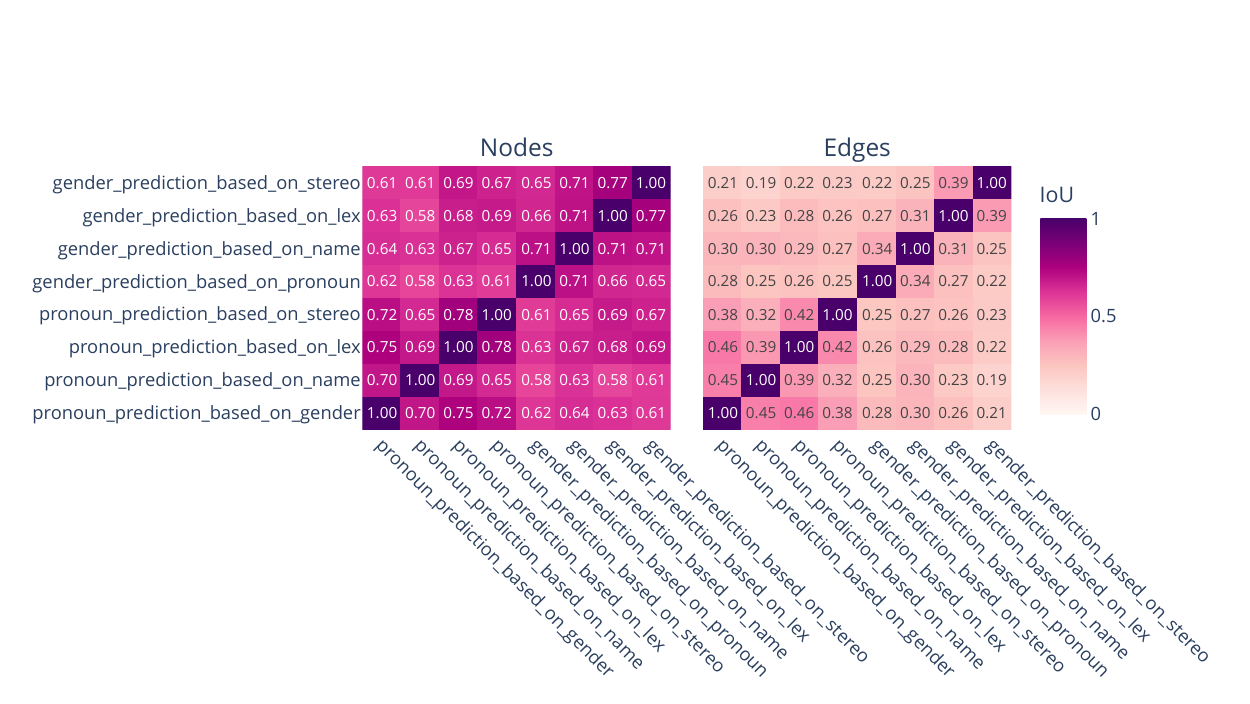}
  \caption{Edge and node intersection over union (Jaccard similarity) for minimal, faithful circuits in Olmo.}
  \label{fig:olmo_iou}
\end{figure*}

\begin{figure*}[]
  \centering
      \centering
    \includegraphics[width=\columnwidth]{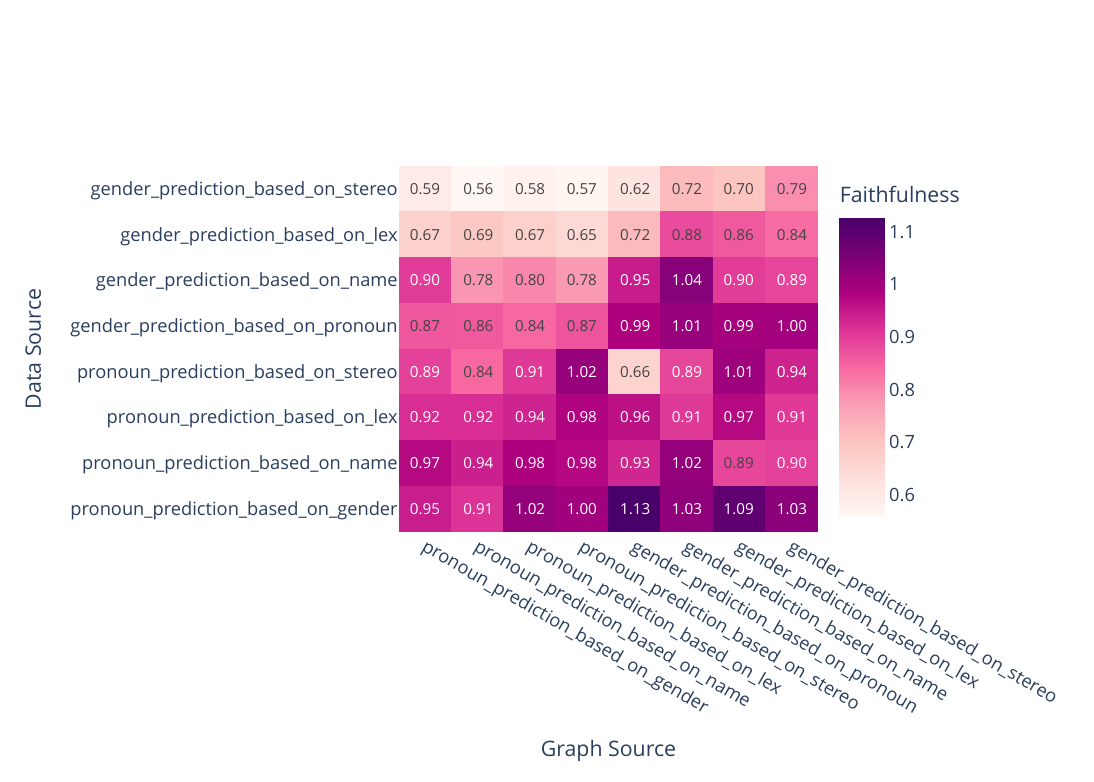}
  \caption{Cross-task faithfulness between the gendered tasks of GKnow, for Olmo.}
  \label{fig:olmo_crosstask_faith}
\end{figure*}

 \Cref{fig:olmo_iou,fig:olmo_crosstask_faith} depict the circuit analysis visualizations for Olmo, omitted from the main document to improve readability and due to lack of space.

\section{Neuron Retrieval and Analysis: Details}
\label{sec:appendixC}

\subsection{Formal Description of Integrated Gradients}
\label{app:Integrated Gradients Formal Description}

Formally, given an input sentence $x$, the model output $P_x(\widehat{w}_i^{l})$ is defined as the probability of the expected output:
\begin{equation}
  P_x(\widehat{w}_i^{(l)}) = p(y^* = x|x, w_i^{(l)} = \widehat{w}_i^{(l)}), \tag{1}
\end{equation}
where $y^*$ is the expected output; $w_i^{(l)}$ is the $i_{th}$ intermediate neuron in the $l$-th layer FFN; and $\widehat{w}_i^{(l)}$ is the constant that $w_i^{(l)}$ is assigned to. To quantify the contribution of a neuron to the prediction, the value of neuron $w_i^{(l)}$ is gradually changed from 0 to its original value $\widehat{w}_i^{(l)}$, and the gradients are integrated. Following the authors, we use a Riemann approximation of the continuous integral, with 20 approximation steps:
\begin{equation}
\text{Attr}(\tilde{w}_i^{(l)}) = \frac{\overline{w}_i^{(l)}}{m} \sum_{k=1}^{m} \frac{\partial P_x(\frac{k}{m}\overline{w}_i^{(l)})}{\partial w_i^{(l)}}
\end{equation}

\begin{figure*}[t]
  \centering

  \begin{subfigure}[t]{0.48\textwidth}
    \centering
    \includegraphics[width=\linewidth]{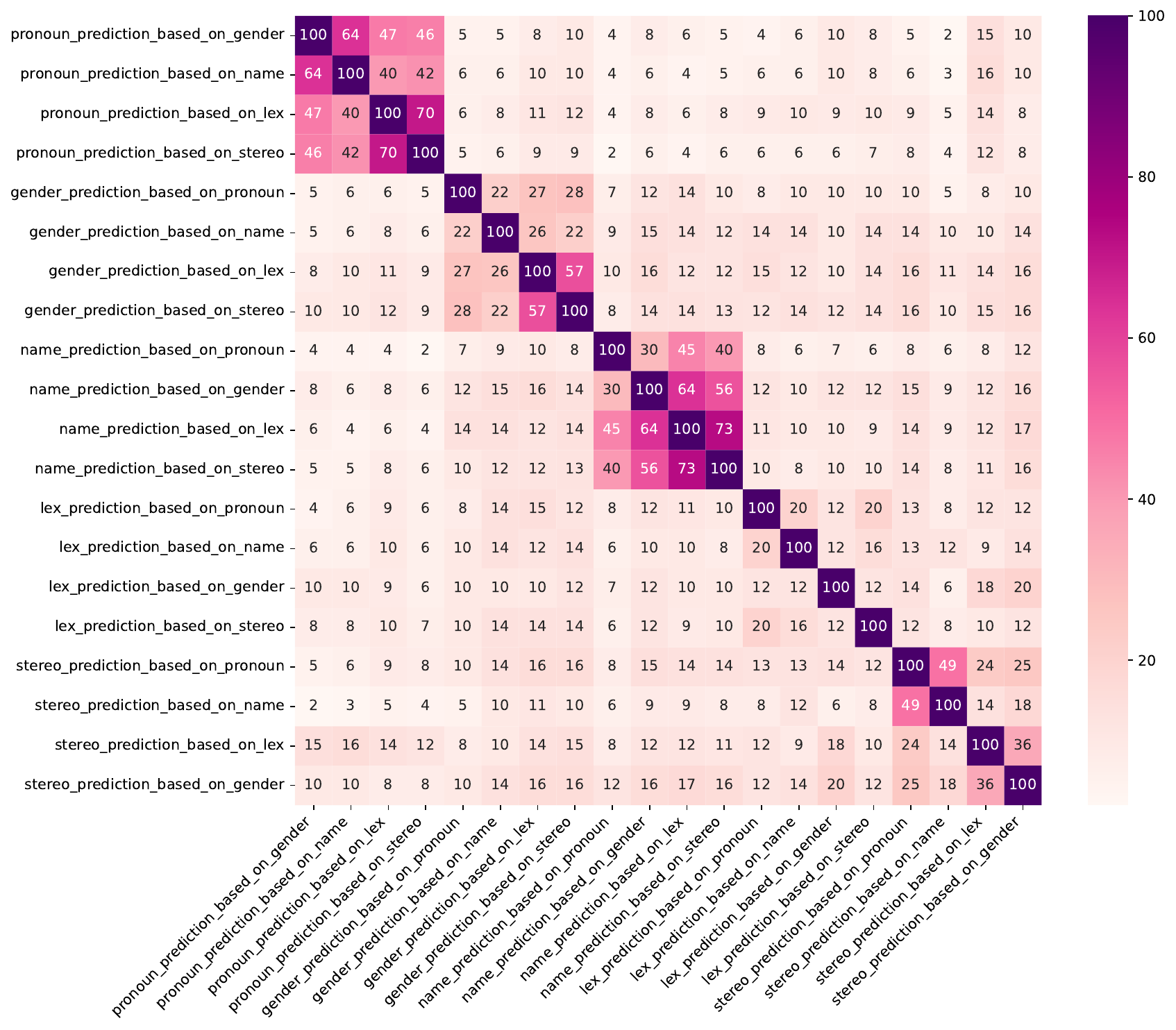}
  \end{subfigure}
  \hfill
  \begin{subfigure}[t]{0.48\textwidth}
    \centering
    \includegraphics[width=\linewidth]{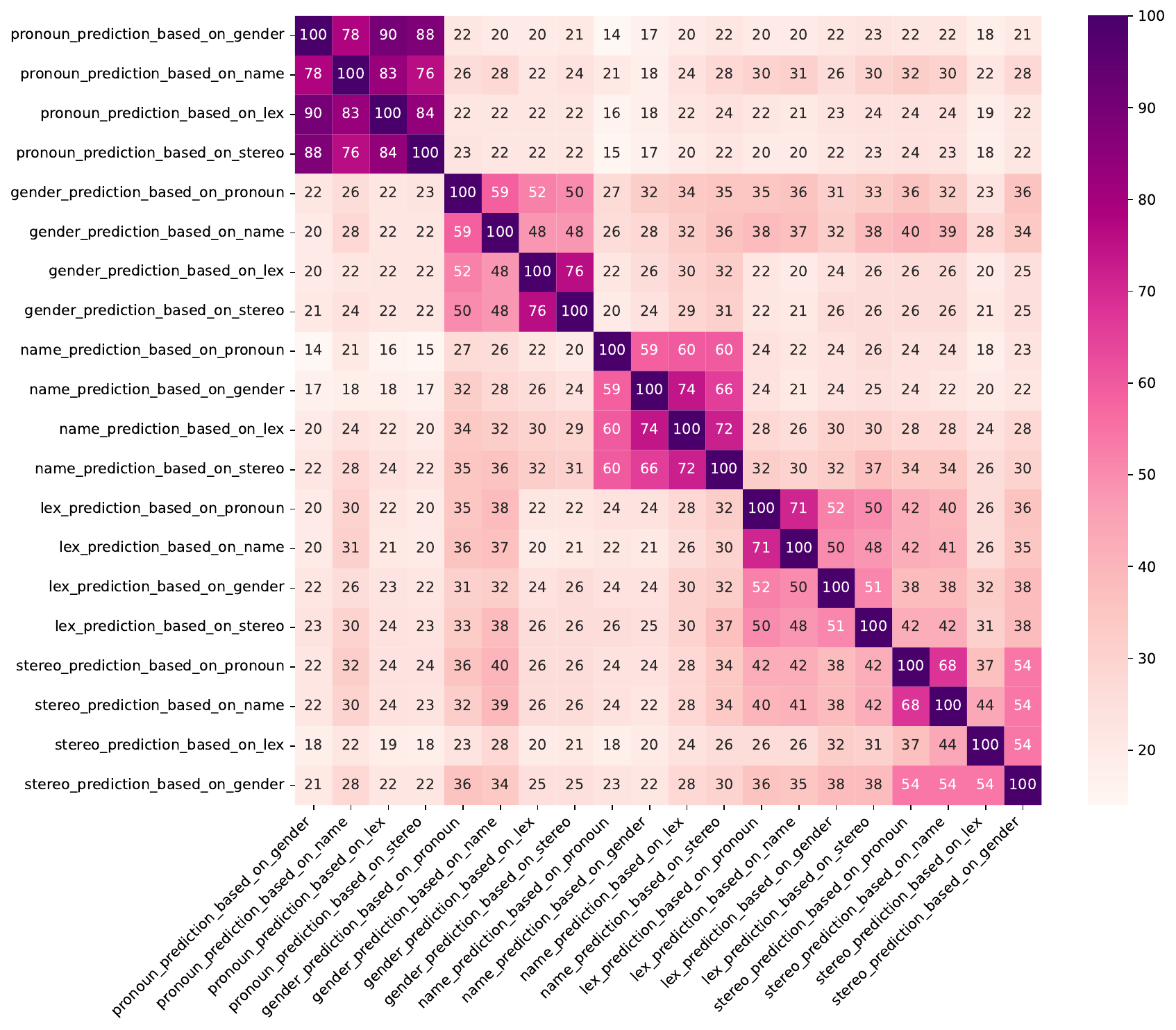}
  \end{subfigure}

  \caption{Overlap of the top 100 IG neurons across \gknow subsets, for Llama (left) and Olmo (right).}
  \label{fig:neuron_overlap}
\end{figure*}

\Cref{fig:neuron_overlap} depicts the overlap across \gknow subsets for the top-100 IG neurons. Overlap is higher within the same type of prediction. Overlap between \texttt{based\_on\_lex} and \texttt{based\_on\_stereo} subsets is also high, which is in line with our circuit analysis observations and supports the hypothesis of entanglement of gender bias and factual gender.





\subsection{Effects of Neuron Ablation on POS-tag Distribution}
\label{test}

We retrieve the top 10 output tokens for the \texttt{pronoun\_prediction\_based\_on\_stereo} subset of \gknow, for Olmo (Table~\ref{tab:before_after_tokens}). The experiment can be reproduced for other subsets, but we reduce the scope here for simplicity. We use \texttt{nltk}'s POS tagger to tag the output tokens' role in the prompt. POS tags with only 1 occurrence, as well as tags referring to punctuation signs, were removed. Ablation causes a decrease in the output probability of personal pronouns, as expected. This change is accompanied by an increase in the output probability of unexpected tokens, such as verbs.

\begin{table}[]
  \centering
  \small
  \begin{tabular}{@{}lrr@{}}
  \toprule
  \textbf{Tag} & \textbf{Before} & \textbf{After} \\
  \midrule
  Personal pronoun & 308 & 130 \\
  Verb, base form & 232 & 344 \\
  Determiner & 75 & 81 \\
  Adverb & 44 & 80 \\
  Adjective & 35 & 48 \\
  Preposition or subordinating conjunction & 32 & 31 \\
  Possessive pronoun & 27 & 27 \\
  Existential there & 21 & 20 \\
  Verb, gerund or present participle & 3 & 16 \\
  Noun, singular or mass & 3 & 3 \\
  \bottomrule
  \end{tabular}
  \caption{Comparison of part-of-speech tag occurrences in the top 10 tokens, before and after the ablation of the top 5 IG neurons in Olmo, for the \texttt{pronoun\_prediction\_based\_on\_stereo} subset of \gknow. Personal pronouns suffer a great decrease in occurrence, while the occurrence of unrelated tokens, such as verbs, increases.}
  \label{tab:before_after_tokens}
\end{table}

\subsection{Neuron Ablation: Additional Results}
\label{app:Neuron Ablation: Additional Results}

Both zero-ablation and mean-ablation are widely used methods to measure feature importance and erase concepts. Since zero-ablation has been more commonly reported in bias-related mechanistic interpretability works, we report those results in the main document. Results for mean-ablation are depicted in \Cref{tab:eval_mean_ablation}. It is worth to note that both zero-ablation and mean-ablation suffer from an out-of-distribution problem, since setting certain activation values to zero or their mean could result in an input that was never observed during training \citep{li2024optimal}. An extensive study of the impact of different ablation methods in biased representations is a possible avenue for future work.

For additional comparison with the main ablation results, we ablate random neurons (\Cref{tab:eval_random_ablation}), impact of which is not statistically significant. Additionally, we select the neurons that are only present in the \texttt{based\_on\_stereo} subset of \texttt{gender\_prediction} prompts, to analyze the impact of ``stereotypical-only'' neurons (\Cref{tab:eval_stereo_ablation}). We conclude that ``stereotypical-only'' neuron ablation is not as impactful as the main ablation results, since it is not enough to change the model's final predictions.

\begin{table*}[]
\centering
\small
\resizebox{\linewidth}{!}{%
\begin{tabular}{p{0.2cm} l c c c c c c c}
\toprule
N & Dataset & $P_{exp}$ & $P_{opp}$ & $P_{other}$ & $\%exp$ & $\%opp$ & $\%other$ & $\Delta_{f,m}$ \\
\midrule
\multirow{5}{*}{0} & GKnow Stereo & 67.66 & 28.90 & 3.43 & 78.75 & 21.25 & 0.00 & 21.81 \\
 & GKnow Factual & 91.49 & 7.17 & 1.34 & 100.00 & 0.00 & 0.00 & 43.77 \\
 & StereoSet & 65.26 & 26.27 & 8.48 & 65.38 & 20.19 & 14.42 & - \\
 & DiFair Neutral & - & - & - & - & - & - & 6.48 \\
 & DiFair Specific & - & - & - & - & - & - & 45.57 \\
 \midrule
\multirow{5}{*}{10} & GKnow Stereo & 66.45 {*} {\scriptsize\textcolor{red!70!black}{$\downarrow$1.21}} & 28.72 {*}{\scriptsize\textcolor{red!70!black}{$\downarrow$0.19}} & 4.83 {\scriptsize\textcolor{green!60!black}{$\uparrow$1.40}} & 82.50 {\scriptsize\textcolor{green!60!black}{$\uparrow$3.75}} & 17.50 {\scriptsize\textcolor{red!70!black}{$\downarrow$3.75}} & 0.00 {\scriptsize\textcolor{gray!70!black}{$\rightarrow$ 0.00}} & 20.53 {\scriptsize\textcolor{red!70!black}{$\downarrow$1.28}}\\
 & GKnow Factual & 91.01 {\scriptsize\textcolor{red!70!black}{$\downarrow$0.48}} & 7.41 {*}{\scriptsize\textcolor{green!60!black}{$\uparrow$0.24}} & 1.57 {\scriptsize\textcolor{green!60!black}{$\uparrow$0.23}} & 100.00 {\scriptsize\textcolor{gray!70!black}{$\rightarrow$ 0.00}} & 0.00 {\scriptsize\textcolor{gray!70!black}{$\rightarrow$ 0.00}} & 0.00 {\scriptsize\textcolor{gray!70!black}{$\rightarrow$ 0.00}} & 42.54 {\scriptsize\textcolor{red!70!black}{$\downarrow$1.23}}\\
 & StereoSet & 65.34 {*} {\scriptsize\textcolor{green!60!black}{$\uparrow$0.08}} & 26.19 {*}{\scriptsize\textcolor{red!70!black}{$\downarrow$0.08}} & 8.48 {*}{\scriptsize\textcolor{gray!70!black}{$\rightarrow$0.00}} & 65.38 {\scriptsize\textcolor{gray!70!black}{$\rightarrow$ 0.00}} & 20.19 {\scriptsize\textcolor{gray!70!black}{$\rightarrow$ 0.00}} & 14.42 {\scriptsize\textcolor{gray!70!black}{$\rightarrow$ 0.00}} \\
 & DiFair Neutral & - & - & - & - & - & - & 6.54 {*}{\scriptsize\textcolor{green!60!black}{$\uparrow$0.06}} \\
 & DiFair Specific & - & - & - & - & - & - & 44.39 {\scriptsize\textcolor{red!70!black}{$\downarrow$1.18}}\\
 \midrule
\multirow{5}{*}{50} & GKnow Stereo & 61.52 {\scriptsize\textcolor{red!70!black}{$\downarrow$6.14}} & 31.16 {\scriptsize\textcolor{green!60!black}{$\uparrow$2.25}} & 7.33 {\scriptsize\textcolor{green!60!black}{$\uparrow$3.89}} & 68.75 {\scriptsize\textcolor{red!70!black}{$\downarrow$10.00}} & 28.75 {\scriptsize\textcolor{green!60!black}{$\uparrow$7.50}} & 2.50 {\scriptsize\textcolor{green!60!black}{$\uparrow$2.50}} & 18.44 {\scriptsize\textcolor{red!70!black}{$\downarrow$3.37}} \\
 & GKnow Factual & 88.20 {\scriptsize\textcolor{red!70!black}{$\downarrow$3.29}} & 9.69 {\scriptsize\textcolor{green!60!black}{$\uparrow$2.51}} & 2.11 {\scriptsize\textcolor{green!60!black}{$\uparrow$0.77}} & 98.90 {\scriptsize\textcolor{red!70!black}{$\downarrow$1.10}} & 1.10 {\scriptsize\textcolor{green!60!black}{$\uparrow$1.10}} & 0.00 {\scriptsize\textcolor{gray!70!black}{$\rightarrow$ 0.00}} & 39.77 {\scriptsize\textcolor{red!70!black}{$\downarrow$6.00}} \\
 & StereoSet & 65.45 {*} {\scriptsize\textcolor{green!60!black}{$\uparrow$0.20}} & 26.03 {*}{\scriptsize\textcolor{red!70!black}{$\downarrow$0.23}} & 8.51  {*}{\scriptsize\textcolor{green!60!black}{$\uparrow$0.04}} & 65.38 {\scriptsize\textcolor{gray!70!black}{$\rightarrow$ 0.00}} & 20.19 {\scriptsize\textcolor{gray!70!black}{$\rightarrow$ 0.00}} & 14.42 {\scriptsize\textcolor{gray!70!black}{$\rightarrow$ 0.00}} & - \\
 & DiFair Neutral & - & - & - & - & - & - & 6.24 {*} {\scriptsize\textcolor{red!70!black}{$\downarrow$0.24}} \\
 & DiFair Specific & - & - & - & - & - & - & 43.22 {\scriptsize\textcolor{red!70!black}{$\downarrow$2.35}} \\
\bottomrule
\toprule
N & Dataset & $P_{exp}$ & $P_{opp}$ & $P_{other}$ & $\%exp$ & $\%opp$ & $\%other$ & $\Delta_{f,m}$ \\
\midrule
\multirow{5}{*}{0} & GKnow Stereo & 63.66 & 26.62 & 9.72 & 77.50 & 16.25 & 6.25 & 17.09 \\
 & GKnow Factual & 90.07 & 7.95 & 1.98 & 98.90 & 1.10 & 0.00 & 40.1 \\
 & StereoSet & 65.55 & 26.63 & 7.82 & 68.27 & 19.23 & 12.50 & - \\
 & DiFair Neutral & - & - & - & - & - & - & 9.63 \\
 & DiFair Specific & - & - & - & - & - & - & 48.26 \\
 \midrule
\multirow{4}{*}{10} & GKnow Stereo & 58.55 {\scriptsize\textcolor{red!70!black}{$\downarrow$5.11}} & 29.01 {*}{\scriptsize\textcolor{green!60!black}{$\uparrow$2.40}} & 12.44 {\scriptsize\textcolor{green!60!black}{$\uparrow$2.72}} & 67.50 {\scriptsize\textcolor{red!70!black}{$\downarrow$10.00}} & 25.00 {\scriptsize\textcolor{green!60!black}{$\uparrow$8.75}} & 7.50 {\scriptsize\textcolor{green!60!black}{$\uparrow$1.25}} & 14.33 {\scriptsize\textcolor{red!70!black}{$\downarrow$2.76}}\\
 & GKnow Factual & 86.83 {\scriptsize\textcolor{red!70!black}{$\downarrow$3.24}} & 10.72 {\scriptsize\textcolor{green!60!black}{$\uparrow$2.77}} & 2.45 {\scriptsize\textcolor{green!60!black}{$\uparrow$0.46}} & 97.80 {\scriptsize\textcolor{red!70!black}{$\downarrow$1.10}} & 2.20 {\scriptsize\textcolor{green!60!black}{$\uparrow$1.10}} & 0.00 {\scriptsize\textcolor{gray!70!black}{$\rightarrow$ 0.00}} & 37.23 {\scriptsize\textcolor{red!70!black}{$\downarrow$2.87}} \\
 & StereoSet & 65.36 {*} {\scriptsize\textcolor{red!70!black}{$\downarrow$0.19}} & 26.66 {*}{\scriptsize\textcolor{green!60!black}{$\uparrow$0.03}} & 7.97 {\scriptsize\textcolor{green!60!black}{$\uparrow$0.15}} & 68.27 {\scriptsize\textcolor{gray!70!black}{$\rightarrow$ 0.00}} & 19.23 {\scriptsize\textcolor{gray!70!black}{$\rightarrow$ 0.00}} & 12.50 {\scriptsize\textcolor{gray!70!black}{$\rightarrow$ 0.00}} & - \\
 & DiFair Neutral & - & - & - & - & - & - & 9.05 {*} {\scriptsize\textcolor{red!70!black}{$\downarrow$0.58}} \\
 & DiFair Specific & - & - & - & - & - & - & 45.92 {\scriptsize\textcolor{red!70!black}{$\downarrow$2.34}} \\
 \midrule
\multirow{5}{*}{50} & GKnow Stereo & 49.77 {\scriptsize\textcolor{red!70!black}{$\downarrow$13.90}} & 32.76 {\scriptsize\textcolor{green!60!black}{$\uparrow$6.14}} & 17.47 {\scriptsize\textcolor{green!60!black}{$\uparrow$7.75}} & 61.25 {\scriptsize\textcolor{red!70!black}{$\downarrow$16.25}} & 25.00 {\scriptsize\textcolor{green!60!black}{$\uparrow$8.75}} & 13.75 {\scriptsize\textcolor{green!60!black}{$\uparrow$7.50}} & 10.07 {\scriptsize\textcolor{red!70!black}{$\downarrow$7.02}} \\
 & GKnow Factual & 57.53 {\scriptsize\textcolor{red!70!black}{$\downarrow$32.54}} & 24.99 {\scriptsize\textcolor{green!60!black}{$\uparrow$17.05}} & 17.48 {\scriptsize\textcolor{green!60!black}{$\uparrow$15.49}} & 73.08 {\scriptsize\textcolor{red!70!black}{$\downarrow$25.82}} & 12.64 {\scriptsize\textcolor{green!60!black}{$\uparrow$11.54}} & 14.29 {\scriptsize\textcolor{green!60!black}{$\uparrow$14.29}} & 4.45 {\scriptsize\textcolor{red!70!black}{$\downarrow$35.65}}  \\
 & StereoSet & 47.99 {\scriptsize\textcolor{red!70!black}{$\downarrow$17.56}} & 31.50 {*} {\scriptsize\textcolor{green!60!black}{$\uparrow$4.88}} & 20.51 {\scriptsize\textcolor{green!60!black}{$\uparrow$12.69}} & 50.96 {\scriptsize\textcolor{red!70!black}{$\downarrow$17.31}} & 27.88 {\scriptsize\textcolor{green!60!black}{$\uparrow$8.65}} & 21.15 {\scriptsize\textcolor{green!60!black}{$\uparrow$8.65}} & - \\
  & DiFair Neutral & - & - & - & - & - & - & 5.83 {\scriptsize\textcolor{red!70!black}{$\downarrow$3.8}} \\
 & DiFair Specific & - & - & - & - & - & - &  14.98 {\scriptsize\textcolor{red!70!black}{$\downarrow$33.28}} \\
\bottomrule
\end{tabular}
}
\caption{Results of mean-ablating the top 10 and 50 IG neurons in Llama (top) and Olmo (bottom). All results except the ones marked with \{*\} are statistically significant ($p$-value $<0.05$, from $t$-score).}
  \label{tab:eval_mean_ablation}
\end{table*}

\begin{table*}[]
\centering
\small
\begin{tabular}{l l c c c}
\toprule
N & Dataset & $P_{exp}$ & $P_{opp}$ & $P_{other}$ \\
\midrule
\multirow{2}{*}{0} & GKnow Stereo & 67.66 & 28.90 & 3.43  \\
 & GKnow Factual & 91.49 & 7.17 & 1.33 \\
\multirow{2}{*}{10} & GKnow Stereo & 67.68 {\scriptsize\textcolor{green!60!black}{$\uparrow$0.02}} & 28.90 {\scriptsize\textcolor{gray!70!black}{$\rightarrow$0.00}} & 3.42 {\scriptsize\textcolor{red!70!black}{$\downarrow$0.01}} \\
 & GKnow Factual & 91.49 {\scriptsize\textcolor{gray!70!black}{$\rightarrow$0.00}} & 7.17 {\scriptsize\textcolor{gray!70!black}{$\rightarrow$0.00}} & 1.33 {\scriptsize\textcolor{gray!70!black}{$\rightarrow$0.00}}\\
\multirow{2}{*}{50} & GKnow Stereo & 67.68 {\scriptsize\textcolor{green!60!black}{$\uparrow$0.02}} & 28.89 {\scriptsize\textcolor{red!70!black}{$\downarrow$0.01}} & 3.43 {\scriptsize\textcolor{gray!70!black}{$\rightarrow$0.00}}  \\
 & GKnow Factual & 91.50 {\scriptsize\textcolor{green!60!black}{$\uparrow$0.01}} & 7.16 {\scriptsize\textcolor{gray!70!black}{$\rightarrow$0.00}} & 1.33 {\scriptsize\textcolor{gray!70!black}{$\rightarrow$0.00}}\\
\bottomrule
\end{tabular}
\caption{Results of zero-ablating random 10 and 50 IG neurons in Llama. Results are not statistically significant.}
  \label{tab:eval_random_ablation}
\end{table*}

\begin{table*}[]
\centering
\small
\begin{tabular}{l l c c c c c c}
\toprule
N & Dataset & $P_{exp}$ & $P_{opp}$ & $P_{other}$ & $\%exp$ & $\%opp$ & $\%other$ \\
\midrule
\multirow{2}{*}{0} & GKnow Stereo & 67.67 & 28.90 & 3.43 & 78.75 & 21.25 & 0.00 \\
 & GKnow Factual & 91.50 & 7.17 & 1.33 & 100.00 & 0.00 & 0.00 \\
\multirow{2}{*}{10} & GKnow Stereo & 67.77 {*} {\scriptsize\textcolor{green!60!black}{$\uparrow$0.10}} & 28.87 {*} {\scriptsize\textcolor{red!70!black}{$\downarrow$0.03}} & 3.36 {\scriptsize\textcolor{red!70!black}{$\downarrow$0.08}} & 78.75 {\scriptsize\textcolor{gray!70!black}{$\rightarrow$ 0.00}} & 21.25 {\scriptsize\textcolor{gray!70!black}{$\rightarrow$ 0.00}} & 0.00 {\scriptsize\textcolor{gray!70!black}{$\rightarrow$ 0.00}} \\
 & GKnow Factual & 91.54 {\scriptsize\textcolor{green!60!black}{$\uparrow$0.04}} & 7.14 {\scriptsize\textcolor{red!70!black}{$\downarrow$0.03}} & 1.32 {\scriptsize\textcolor{red!70!black}{$\downarrow$0.02}} & 100.00 {\scriptsize\textcolor{gray!70!black}{$\rightarrow$ 0.00}} & 0.00 {\scriptsize\textcolor{gray!70!black}{$\rightarrow$ 0.00}} & 0.00 {\scriptsize\textcolor{gray!70!black}{$\rightarrow$ 0.00}} \\
\multirow{2}{*}{50} & GKnow Stereo & 67.96 {\scriptsize\textcolor{green!60!black}{$\uparrow$0.30}} & 28.67 {\scriptsize\textcolor{red!70!black}{$\downarrow$0.23}} & 3.37 {*} {\scriptsize\textcolor{red!70!black}{$\downarrow$0.07}} & 78.75 {\scriptsize\textcolor{gray!70!black}{$\rightarrow$ 0.00}} & 21.25 {\scriptsize\textcolor{gray!70!black}{$\rightarrow$ 0.00}} & 0.00 {\scriptsize\textcolor{gray!70!black}{$\rightarrow$ 0.00}} \\
 & GKnow Factual & 91.78 {\scriptsize\textcolor{green!60!black}{$\uparrow$0.29}} & 6.90 {\scriptsize\textcolor{red!70!black}{$\downarrow$0.27}} & 1.32 {\scriptsize\textcolor{red!70!black}{$\downarrow$0.02}} & 100.00 {\scriptsize\textcolor{gray!70!black}{$\rightarrow$ 0.00}} & 0.00 {\scriptsize\textcolor{gray!70!black}{$\rightarrow$ 0.00}} & 0.00 {\scriptsize\textcolor{gray!70!black}{$\rightarrow$ 0.00}} \\
\bottomrule
\end{tabular}
\caption{Results of zero-ablating the top 10 and 50 IG neurons in Llama, that are only present in the stereotypical set of neurons (identified with the \gknow set \texttt{gender\_prediction\_based\_on\_stereo}). Changes in probability distribution can be statistically significant, but are not enough to flip the model's predictions.}
  \label{tab:eval_stereo_ablation}
\end{table*}

\begin{table*}[]
   \centering
   \small
   \begin{tabular}{l l p{5cm} p{3cm} p{3cm}}
   \toprule
   \textbf{Model} & \textbf{Neuron} & \textbf{Subsets} & \textbf{Top Tokens} & \textbf{Bottom Tokens} \\
   \midrule
   \multirow{5}{*}{Olmo} 
   & L31N8077 & lex\_prediction\_based\_on\_name, pronoun\_prediction\_based\_on\_gender, pronoun\_prediction\_based\_on\_lex, pronoun\_prediction\_based\_on\_name, pronoun\_prediction\_based\_on\_stereo & \texttt{['spokesman', 'ils', 'handsome', 'ico', 'Brothers']} & \texttt{['she', 'her', 'herself', 'She', 'woman']} \\ 
   & L29N6458 & gender\_prediction\_based\_on\_lex, gender\_prediction\_based\_on\_name, pronoun\_prediction\_based\_on\_lex, pronoun\_prediction\_based\_on\_name, pronoun\_prediction\_based\_on\_stereo & \texttt{['her', 'she', 'hers', 'herself', 'she']} & \texttt{['he', 'his', 'himself', 'his', 'His']} \\
   & L30N10936 & name\_prediction\_based\_on\_gender, name\_prediction\_based\_on\_lex, name\_prediction\_based\_on\_pronoun, name\_prediction\_based\_on\_stereo & \texttt{['Rob', 'Tim', 'Mark', 'Tim', 'Core']} & \texttt{['Laura', 'Sarah', 'Rachel', 'Anna', 'Michelle']} \\
   & L28N8701 & pronoun\_prediction\_based\_on\_gender, pronoun\_prediction\_based\_on\_lex, pronoun\_prediction\_based\_on\_name, pronoun\_prediction\_based\_on\_stereo & \texttt{['they', 'he', 'she', 'They', 'they']} & \texttt{['his', 'don', 'did', 'everyone', 'our']} \\
   & L30N5440 & pronoun\_prediction\_based\_on\_gender, pronoun\_prediction\_based\_on\_lex, pronoun\_prediction\_based\_on\_name, pronoun\_prediction\_based\_on\_stereo & \texttt{['him', 'them', 'Him', 'her', 'him']} & \texttt{['s', 'he', 'She', 'she', 'HE']} \\
   \midrule
   \midrule
   \multirow{5}{*}{Llama} 
   & L23N13431 & name\_prediction\_based\_on\_gender, name\_prediction\_based\_on\_lex, name\_prediction\_based\_on\_pronoun, name\_prediction\_based\_on\_stereo & \texttt{['bey', 'Bey', 'Desc', 'Crom', 'Kop']} & \texttt{['Mark', 'Gene', 'Mark', 'Rob', 'Phil']} \\
   & L24N12384 & name\_prediction\_based\_on\_gender, name\_prediction\_based\_on\_lex, name\_prediction\_based\_on\_pronoun, name\_prediction\_based\_on\_stereo & \texttt{['Ell', 'Eld', 'zier', 'Al', 'Ker']} & \texttt{['John', 'John', 'Jones', 'Smith', 'Johns']} \\
   & L30N6390 & pronoun\_prediction\_based\_on\_gender, pronoun\_prediction\_based\_on\_lex, pronoun\_prediction\_based\_on\_name, pronoun\_prediction\_based\_on\_stereo & \texttt{['his', 'his', 'himself', 'zijn', 'jeho']} & \texttt{['He', 'He', '.He', 'he', '\_he']} \\
   & L30N13342 & pronoun\_prediction\_based\_on\_gender, pronoun\_prediction\_based\_on\_lex, pronoun\_prediction\_based\_on\_name, pronoun\_prediction\_based\_on\_stereo & \texttt{['HIM', 'ihn', 'lád', 'him', 'ihm']} & \texttt{['he', 'He', 'He', 'he', '.he']} \\
   & L28N2183 & pronoun\_prediction\_based\_on\_gender, pronoun\_prediction\_based\_on\_lex, pronoun\_prediction\_based\_on\_name, pronoun\_prediction\_based\_on\_stereo & \texttt{['Him', 'him', 'HIM', 'him', 'ihn']} & \texttt{['he', 'he', '.he', 'He', 'He']} \\
   \bottomrule
   \end{tabular}
   \caption{Interpretable neurons common to several subsets of \gknow, for Olmo and Llama. Notably, some Olmo neurons promote/suppress different types of gender-related tokens: For example, L31N8077 suppresses female pronouns and the noun \textit{woman}.}
   \label{tab:tokens}
\end{table*}

\section{Implementation Details}
All used datasets are in English. We utilized AI assistants to enhance the aesthetic quality and readability of data visualizations. We used a NVIDIA RTX A6000 and a NVIDIA A100-SXM4-80GB GPU to infer Llama-3.1-8B and Olmo-7B (with default parameters), and to identify the neuron circuits described in this work. Replication of all our analyses and experiments takes approximately 6 hours for both models on an RTX A6000.

\end{document}